\pdfoutput=1

\documentclass[11pt]{article}

\usepackage[preprint]{acl}

\usepackage{times} 
\usepackage{latexsym}

\usepackage[T1]{fontenc}

\usepackage[utf8]{inputenc}

\usepackage{microtype}

\usepackage{inconsolata}

\usepackage{graphicx}

\usepackage{amsmath}

\usepackage{booktabs}       
\usepackage{amsfonts}       
\usepackage{nicefrac}       
\usepackage{cleveref}       
\usepackage{subcaption}
\usepackage{float}
\usepackage{tcolorbox}
\usepackage{placeins}
\usepackage{relsize}
\usepackage{wrapfig}
\usepackage{hyperref}

\usepackage{tabularx}

\usepackage{listings}
\lstdefinestyle{omsearchprompt}{
	basicstyle=\ttfamily\scriptsize,
	breaklines=true,
	breakatwhitespace=false,
	columns=fullflexible,
	keepspaces=true,
	showstringspaces=false
}

\usepackage{algorithm}
\usepackage{algpseudocode}

\newcounter{textbox}

\DeclareRobustCommand{\xtag}[1]{\texttt{\char`\<#1\char`\>}}

\hypersetup{
	pdftitle={Output-Space Search: Targeting LLM Generations in a Frozen Encoder-Defined Output Space},
	pdfsubject={cs.CL,cs.AI},
	pdfauthor={Tobias Materzok},
	pdfkeywords={LLMs, controllable generation, output-space search, reinforcement learning, Bayesian optimization},
}

\author{Tobias Materzok \\
	\texttt{t.materzok@theo.chemie.tu-darmstadt.de}
}

\title{Output-Space Search: Targeting LLM Generations in a Frozen Encoder-Defined Output Space}

\begin{document}
	\maketitle

\footnotetext{
	Code (partial release: $Z$-space construction + code-domain training): \url{https://github.com/TobiasMaterzok/OS-Search}.
	Story-domain training code/data are intentionally not released; see \S\ref{sec:limitations}.
}

\begin{abstract}
	\noindent
	We introduce \emph{Output-Space Search} (OS-Search), which turns LLM generation into endpoint search. An outer loop selects a target $z^\ast$ in a frozen encoder-defined 3D output space $Z$, and a retrieval-grounded policy trained with sequence-level RL generates outputs whose coordinates land near $z^\ast$ under standard autoregressive decoding. This enables parallel sweeps and black-box optimization in $Z$ without path-dependent token/program search. On stories, sweeping $Z_{\text{text}}$ yields $3.1\times$ higher LLM-scored diversity than prompt-chaining. On code, Bayesian optimization over $Z_{\text{code}}$ improves an objective withheld from the controller under matched inference budgets while preserving validity.
\end{abstract}

\section{Introduction}\label{introduction}

Large language models (LLMs) are now strong generators of text and code, but many
workflows require more than ``one good sample.'' Users often want to explore
multiple plausible completions in parallel (diverse drafts, branches, alternative
solutions), or search for outputs that maximize a downstream score
available only through an external evaluator (e.g., heuristics,
or learned judges). In practice this is done by repeated sampling in token space or adaptive loops
that feed earlier generations back into the context (e.g., prompt chaining or
rejection/reranking). These can introduce sequential dependence and/or growing
context length, and they still provide no stable low-dimensional interface that
an outer loop can optimize directly.
\begin{figure}[!hb] 
	\centering
	\includegraphics[width=0.82\linewidth,trim=0.05cm 0.4cm 0.1cm 0.06cm,clip]{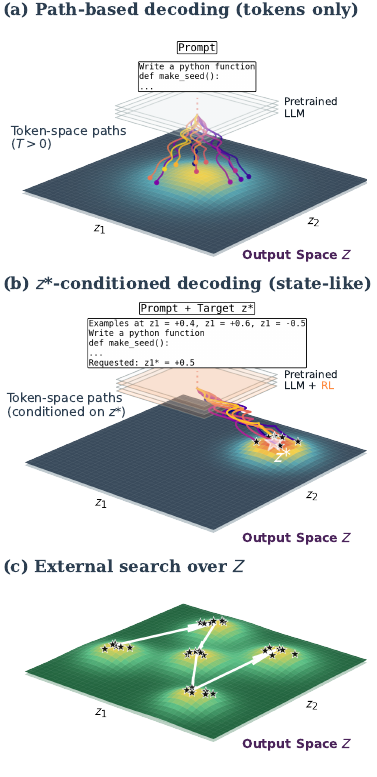}
	\caption[Path- vs state-like control.]{\textbf{Path- vs state-like control (schematic 2D slice of $\mathbf{Z}$).} (a) Standard decoding: a fixed prompt yields token-space paths whose endpoints concentrate in one region of the output space $Z$. (b) $z^\ast$-conditioned decoding: the same base prompt plus a requested target $z^\ast$ (grounded by retrieved exemplars) shifts endpoints toward $z^\ast$ in $Z$ (up to targeting error), while the decoder remains autoregressive. (c) External search: an optimizer moves $z^\ast$ across $Z$ to explore regions and maximize a score $f(x)$. Because branches are conditionally independent given the instantiated prompt, multi-branch generation is parallel.}
	\label{fig:fig_1}
\end{figure}
We ask whether an autoregressive LLM can expose a simple state-like target
that selects where an output should land. Concretely, an outer loop chooses
a target point $z^\ast$ once per sample, the model generates $x$ whose realised
coordinates $z(x)$ lie near $z^\ast$, and the outer loop can sweep or optimize
$z^\ast$ while keeping decoding standard and autoregressive (Fig.~\ref{fig:fig_1}).
We call this control state-like (in the sense of state vs.\ path functions, where the objective depends on the endpoint rather than the full trajectory) because $z^\ast$ specifies an intended endpoint region in $Z$. The outer loop searches over targets $z^\ast$ rather than token paths or program edits.

We implement this interface with \emph{Output-Space Search} (OS-Search), which keeps representation and actuation modular. 
\newline\noindent\textbf{Representation.} We define fixed coordinates $z(x)$ for task outputs using a frozen encoder $E$ and a frozen linear projection (Sec.~\ref{sec:method}, App.~\ref{app:zspace-details}).
\newline\noindent\textbf{Actuation.} We train a retrieval-grounded policy conditioned on a requested target $z^\ast$ so that, via sequence-level RL, its generations land near that request in $Z$ while also self-reporting $\hat z$.
\newline\noindent\textbf{Use.} The resulting controller exposes a black-box actuator $\mathcal{F}$ that maps $(p,z^\ast)$ to a sample $x$, enabling parallel sweeps and black-box optimization over targets in $Z$ without decoding-time guidance.

We evaluate \emph{OS-Search} on both stories (parallel diversity sweeps) and code (outer-loop black-box optimization in $Z$), and report target-tracking and calibration diagnostics.

\paragraph{Contributions.}
We make three contributions.
\begin{itemize}
	\item We propose \emph{OS-Search}: a fixed, external, encoder-defined output space $Z$ together with a $z^\ast$-conditioned controller trained with sequence-level RL to (approximately) hit requested coordinates and self-report $\hat z_S$.
	\item For stories, we build an anchored $Z_{\text{text}}$ and show that anchoring yields a stable reference axis $z_1$, which we validate via a monotonic relationship between requested $z_1^\ast$ and an automated templatedness proxy (EQ-Bench Slop-Score). Sweeping a small grid in $Z_{\text{text}}$ yields large embarrassingly-parallel multi-branch diversity gains.
	\item For code, we build an unanchored $Z_{\text{code}}$ and show it supports black-box optimization: Bayesian optimization (BO) over $z^\ast$ improves a withheld executable objective under matched valid-program budgets.
\end{itemize}

\noindent Tab.~\ref{tab:capability_comparison} contrasts \emph{OS-Search} with direct search methods (e.g., best-of-$N$, MCTS, evolutionary program search) at the level of the control/search interface.

\begin{table}[t]
	\centering
	\small
	\renewcommand{\arraystretch}{1.1}
	\begin{tabular}{p{0.44\columnwidth} p{0.44\columnwidth}}
		\toprule
		\textbf{Trajectory/program search} & \textbf{OS-Search} \\
		\midrule
		searches token paths / program edits & searches state-like targets $z^\ast$ \\
		high-dim, discrete search & low-dim, continuous search ($d_z{=}3$) \\
		evaluator drives the inner loop & evaluator used only in outer loop, can be withheld \\
		learns objective-specific solver/heuristic & learns an actuator $(p,z^\ast)\!\mapsto\!x$ without access to the downstream objective \\
		sequential dependence across steps & independent samples per $z^\ast$ (parallel sweeps) \\
		best-of-run artifact & reusable coordinate interface $Z$ \\
		\bottomrule
	\end{tabular}
	\caption{Capabilities and typical behaviors of \emph{OS-Search} versus direct trajectory/program search.}
	\label{tab:capability_comparison}
\end{table}

\section{Related Work}\label{sec:related_work}

\paragraph{Control for LLMs via prompts, latents, and decoding-time guidance.}
Our goal is a low-dimensional, user-facing control space that an outer loop can sweep or optimize over task outputs while keeping decoding standard and autoregressive.
Many widely used levers are path-based.
They modify the token-level sampling process (e.g., temperature, top-$p$, repetition penalties, logit bias) or optimize preferences over complete trajectories (e.g., RLHF-style objectives)~\citep{christiano_deep_2023,shao_deepseekmath_2024,deepseek-ai_deepseek-r1_2025,zheng_group_2025,openai_openai_2024}.
Most controllable-generation methods instead inject control variables into the prompt, the model, or the sampling loop.
This includes discrete control codes/tokens~\citep{keskar_ctrl_2019}, learned continuous prompts/prefixes/adapters~\citep{lester_power_2021,li_prefix-tuning_2021,houlsby_parameter-efficient_2019}, internal latent-variable control in text VAEs/flows~\citep{gu_controllable_2023,ding_maclasa_2023,abeer_multi-objective_2024,shi_uniform_2024}, and activation-space steering via steering directions or features~\citep{subramani_extracting_2022,liang_controllable_2024,turner_steering_2024,arditi_refusal_2024}, including interventions that amplify sparse autoencoder (SAE) features~\citep{cunningham_sparse_2023,gao_scaling_2024}.
Decoding-time guidance such as PPLM and GeDi recomputes attribute signals on the prefix and modifies logits online~\citep{dathathri_plug_2020,krause_gedi_2020}, tightly coupling control to the sampled trajectory.

Treating generation as optimization over a continuous representation space is common in inverse design, e.g., mapping discrete molecules to a continuous latent space and optimizing objectives in that space before decoding back to valid structures~\citep{gomez-bombarelli_automatic_2018}.

These mechanisms can be effective, but their control variables are often model-internal, high-dimensional, or prefix/path-coupled (and may shift across model updates).
AxBench finds current steering methods lag strong prompting/fine-tuning baselines for concept control, with SAE-based steering not competitive in their evaluation~\citep{wu_axbench_2025}.
\emph{OS-Search} can be viewed as conditional generation with a continuous condition, but it differs in where the condition lives and how it is used.
We define external encoder-defined coordinates $z(x)$ for task outputs, make target requests actionable by pairing each $z^\ast$ with nearby exemplars in $Z$, and train a policy to follow these retrieval-grounded requests (with calibration and $\mathrm{Success}@\varepsilon$ diagnostics).
This exposes a fixed coordinate interface $Z$ that an outer loop can sweep or optimize once per sample without decoding-time guidance.

\paragraph{Search-based agents and algorithm discovery.}
Most LLM-based discovery systems perform trajectory/edit search: they iteratively propose token-level actions or program edits, evaluate candidates, and use the evaluator to drive the inner loop (e.g., MCTS, evolutionary program search, or test-time RL)~\citep{silver_mastering_2017,silver_mastering_2016,chen_alphamath_2024,xie_monte_2024,feng_alphazero-like_2024,romera-paredes_mathematical_2024,novikov_alphaevolve_2025,yuksekgonul_learning_2026}.
\emph{OS-Search} instead aims to make search endpoint-based: we train an objective-agnostic actuator that follows requested coordinates in a frozen output space $Z$, and use downstream evaluators only in an outer loop that searches over $z^\ast$ (with $f(x)$ withheld from the controller).
A growing line of work pairs strong function approximators with explicit search and automated evaluation to discover policies and algorithms: AlphaZero-style agents combine self-play with Monte Carlo tree search to discover superhuman strategies~\citep{silver_mastering_2017,silver_mastering_2016,chen_alphamath_2024,xie_monte_2024,feng_alphazero-like_2024}, and related systems have been applied to algorithmic domains such as tensor decomposition and matrix multiplication~\citep{fawzi_discovering_2022}.
More recently, FunSearch and AlphaEvolve use LLM-guided evolutionary search over programs, paired with automated evaluators, to discover new mathematical constructions and optimized heuristics~\citep{romera-paredes_mathematical_2024,novikov_alphaevolve_2025}.
Concurrently, TTT-Discover performs reinforcement learning at test time to adapt an LLM on a single problem instance using continuous evaluator rewards, explicitly prioritizing discovery of one best-of-run solution over average-case performance~\citep{yuksekgonul_learning_2026}.

\paragraph{Retrieval grounding for numeric targets in $Z$.}
Retrieval-augmented generation (RAG) retrieves documents to inject world knowledge~\citep{lewis_retrieval-augmented_2020}. In \emph{OS-Search}, retrieval plays a different role. It grounds otherwise-arbitrary numeric coordinates by retrieving nearby outputs in $Z$ (plus a contrast example), making a requested target $z^\ast$ interpretable and actionable for the generator (Fig.~\ref{fig:prompt-schema}).

\paragraph{Multi-branch story generation.}
A parallel line of work explicitly constructs branching story trees from
a single prompt, either by expanding a latent story graph or by sampling
choice points and continuations~\citep{alabdulkarim_goal-directed_2021,wen_grove_2023,nottingham_improving_2024,materzok_cosmos_2025}.
These methods focus on the structure of branching narratives and
typically rely on sampling and search over discrete branches.
Avoidance Decoding~\citep{park_avoidance_2025} introduces a decoding-time
similarity penalty that discourages new branches from resembling earlier ones,
and demonstrates strong gains in multi-branch diversity.
Our approach is complementary. Rather than applying guidance at each decoding
step, we expose a low-dimensional external output space that can be swept or
optimized once per sample via a state-like target $z^\ast$.

\section{Methods}\label{sec:method}

We define the frozen coordinate map and the sequence-level RL objective for training a $z^\ast$-conditioned controller. Alg.~\ref{alg:os-search} summarises the full pipeline, including target sweeps and (when an evaluator is available) outer-loop optimization in $Z$.

\paragraph{Notation and setup.}
Let $x$ denote the task content (the generated \xtag{text} field) and $z(x)\in Z\subseteq\mathbb{R}^{d_z}$ its frozen coordinates under a fixed encoder+projection (PCA subspace + Varimax rotation~\citep{kaiser_varimax_1958}). Each instance specifies a base prompt $p$, a requested target $z^\ast$, and constrained axes $S\subseteq\{1,\dots,d_z\}$. We sample a structured completion $y$ containing \xtag{text} $x$ and a self-report $\hat z_S$ (Fig.~\ref{fig:prompt-schema}), and write $z := z(x)$. Only axes in $S$ enter the reward, and $|S|$ is the control dimensionality. In all our experiments $d_z=3$.

\begin{figure}[htb]
	\centering
	\fbox{%
		\begin{minipage}{0.95\linewidth}
			\small\ttfamily
			USER:\\
			\quad Guidance examples (retrieved in $Z$):\\
			\quad\quad [\xtag{text} x\_1 \xtag{/text}\xtag{target} $z(x\_1)$ \xtag{/target}] \ (near $z^\ast$)\\
			\quad\quad [\xtag{text} x\_2 \xtag{/text}\xtag{target} $z(x\_2)$ \xtag{/target}] \ (near $z^\ast$)\\
			\quad\quad [\xtag{text} x\_3 \xtag{/text}\xtag{target} $z(x\_3)$ \xtag{/target}] \ (near $-z^\ast$)\\
			\quad Task + domain constraints\\
			\quad REQUESTED TARGET: $z^\ast_S$ (e.g.\ z1=+0.5, z2=+0.2, z3=+0.2)\\[0.25em]
			ASSISTANT:\\
			\quad \xtag{think} ... \xtag{/think}\\
			\quad \xtag{title} ... \xtag{/title}\\
			\quad \xtag{text} new program/story $x$ \xtag{/text}\\
			\quad \xtag{target} $\hat z_S$ \xtag{/target}
		\end{minipage}%
	}
	\caption{\textbf{Prompt schema for OS-Search.}
		The prompt includes retrieved exemplars with stored realised coordinates $z(x_j)$ and the numeric request $z^\ast_S$.
		The model outputs a structured completion $y$ containing task content $x$ and a self-report $\hat z_S$.
		We compute realised coordinates by embedding only \xtag{text} to obtain $z(x)$.}
	\label{fig:prompt-schema}
\end{figure}

\begin{algorithm}[htb]
	\caption[OS-Search: fixed output coordinates + state-like target search in Z.]{\textbf{OS-Search: fixed output coordinates + state-like target search in $Z$.}}
	\label{alg:os-search}
	\footnotesize
	\begin{algorithmic}
		\Require Frozen encoder $E(\cdot)$, corpus $\mathcal{D}$, (optional) anchors $\mathcal{A}$, $d_z{=}3$.
		
		\State \textbf{(A) Build $Z$ and the retrieval library.}
		\State Fit $\mu$ and $U$ (PCA$\rightarrow$Varimax, optionally anchor $z_1$ with $\mathcal{A}$).
		\State Define $z(x)=U^\top(E(x)-\mu)$.
		\State Build exemplar library $\mathcal{L}=\{(x,z(x))\}$ and a nearest neighbor (NN) index in $Z$ (for code, we optionally grow $\mathcal{L}$ during RL with valid model-generated programs).
		\State Freeze $(E,\mu,U)$.
		
		\State \textbf{(B) Train a $z^\ast$-conditioned controller $\pi_\theta(\cdot\mid p,z^\ast)$.}
		\For{group-based RL updates}
		\State Sample base prompt $p$, constrained axes $S$, and target $z^\ast$ (curriculum).
		\State Retrieve exemplars from $\mathcal{L}$: two near $z^\ast$ and one near $-z^\ast$, instantiate the prompt as in Fig.~\ref{fig:prompt-schema}.
		\State Sample a group of structured completions from $\pi_\theta(\cdot\mid p,z^\ast)$, each containing \xtag{text} and \xtag{target} ($\hat z_S$).
		\For{each completion $y=(x,\hat z_S)$ in the group}
		\State $R_{\text{format}} \gets \textsc{Valid}(y)\in\{0,1\}$. \Comment{Parse + domain gate: text coherence, code AST+execute+board}
		\State If $R_{\text{format}}=0$, set $R=0$. Else embed only \xtag{text} to get $z(x)$ and compute
		\State $R = 3 + 3R_{\text{dist}}(z(x),z^\ast;S) + 1.5R_{\text{hon}}(\hat z,z(x);S)$ (Sec.~\ref{subsec:zspace-rl}).
		\EndFor
		\State Update $\theta$ with group-based RL using group-relative rewards.
		\EndFor
		
		\State \textbf{(C) Use: sweep or optimize targets in $Z$ (optional evaluator $f$).}
		\State Fix a base prompt $p$.
		\State Propose targets $z^\ast$ by grid/random/BO.
		\For{each proposed $z^\ast$}
		\State Retrieve exemplars and instantiate the prompt as in Fig.~\ref{fig:prompt-schema}.
		\State Sample $K$ candidates $x_k\sim\pi_\theta(\cdot\mid p,z^\ast)$ and optionally select $x_{\text{best}}$ (e.g., maximize $f$).
		\State If using BO, update the surrogate at realised coordinates $(z(x_{\text{best}}), f(x_{\text{best}}))$.
		\EndFor
	\end{algorithmic}
\end{algorithm}

\subsection{Representation: constructing the output space for text and code}
\label{subsec:latent-z}

We construct a low-dimensional output space $Z$ (a coordinate system over task outputs) on top of a frozen encoder. Given task content $x$ (the generated \xtag{text} block, for code we extract the \texttt{make\_seed()} function and strip comments before embedding, App.~\ref{app:prompt-format}), the encoder produces an embedding $E(x)\in\mathbb{R}^D$, and a frozen linear map projects it to coordinates
\[
z(x) \;=\; U^\top(E(x)-\mu) \in \mathbb{R}^{d_z},
\]
where $\mu$ is the corpus mean embedding and $U\in\mathbb{R}^{D\times d_z}$ has orthonormal columns. We fit $(\mu,U)$ once per domain (PCA followed by an orthogonal Varimax rotation within the PCA subspace, additional anchoring for our story experiments) and then freeze $E$, $U$, and $\mu$ throughout RL so training changes only the policy.

For stories, we fit $Z_{\text{text}}$ on a story corpus and anchor $z_1$ toward Qwen3-generated ``default-style'' stories relative to the corpus mean (App.~\ref{app:zspace-details}). For code, we fit $Z_{\text{code}}$ on a library of $N{=}188$ valid \texttt{make\_seed()} implementations (used to fit $Z_{\text{code}}$, for the code warm-start stage, App.~\ref{app:curriculum}, and as the Baseline ($Z_{\text{code}}$) sweep in Sec.~\ref{sec:code-results}). App.~\ref{app:zspace-details} provides full construction details (including the motivation for a task-specific code space under our \texttt{make\_seed} constraints), diagnostics, and hyperparameter selection.

\subsection{Actuation: training a \texorpdfstring{$z^\ast$}{z-star}-conditioned controller}
\label{subsec:zspace-rl}

\paragraph{Goal.}
Given a base prompt $p$ and a requested target $z^\ast$ (optionally only on a subset of axes $S$),
we train a controller that samples a structured completion $y$ whose task content
$x$ lands near $z^\ast$ in the frozen output space $Z$.

\paragraph{Prompt interface and required output.}
Requests follow the schema in Fig.~\ref{fig:prompt-schema}. We retrieve exemplars in $Z$ (two near $z^\ast$, one near $-z^\ast$) and include them together with the numeric request $z^\ast_S$. The model outputs a structured completion $y$ containing task content $x$ in \xtag{text} and a self-report $\hat z_S$ in \xtag{target}. For scoring we embed only the generated \xtag{text} block to obtain $z(x)$. We also request a brief \xtag{think} hypothesis (not embedded), which should help the policy verbalize how it intends to realize the target.
\begin{figure*}[htb]
	\centering
	\includegraphics[width=0.7\linewidth]{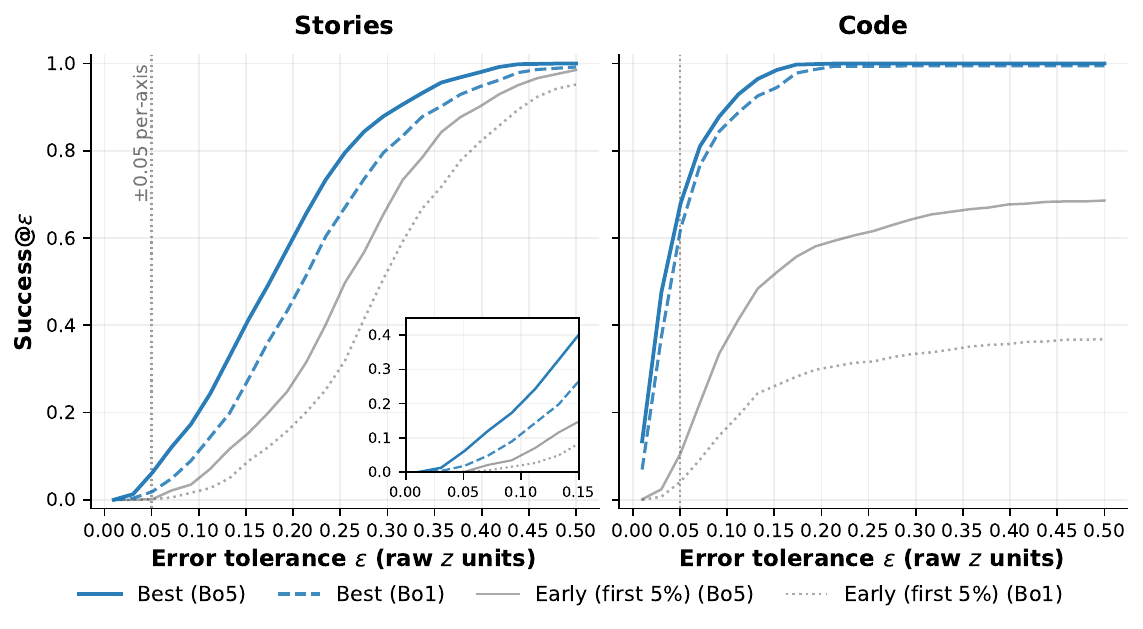}
	\caption[Success--tolerance curves for 3D control in Z.]{
		\textbf{Success--tolerance curves for 3D control in $Z$.}
		Dashed curves show one-shot sampling, and solid curves show best-of-5 at the same request.
		Malformed/invalid outputs count as failures. Best checkpoint (blue) against first 5\% of the RL run (gray).}
	\label{fig:success_tolerance}
\end{figure*}

\paragraph{Why exemplars ground numeric targets.}
Because $Z$ is defined by a frozen encoder plus an arbitrary rotation within a PCA subspace,
the meaning of numeric coordinates is not self-evident.
Nearest-neighbour exemplars provide local grounding. They show what outputs typically look like
around the requested neighborhood in $Z$.
To prevent trivial copying in code, near-duplicate outputs to retrieved exemplars receive zero reward during RL (App.~\ref{app:rl-config}).

\paragraph{Sequence-level reward for validity, targeting, and honesty.}
For each sampled structured completion $y \sim \pi_\theta(\cdot \mid p, z^\ast)$,
we parse out $x$ and $\hat z_S$ and compute a scalar reward
\[
\begin{aligned}
	R(y) = R_{\text{format}}(y)\Bigl(&3 + 3R_{\text{dist}}(z(x), z^\ast; S) \\
	&\quad + 1.5R_{\text{hon}}(\hat z, z(x); S)\Bigr).
\end{aligned}
\]
Here, $R_{\text{format}}(y)\in\{0,1\}$ is a hard parse + domain-validity gate.
If it fails, we set $R(y)=0$ and do not score distance or honesty.
For code, this gate includes AST sanitization and execution to a valid $16\times16$ board.
$R_{\text{dist}}$ rewards targeting: it increases as the realised coordinate $z(x)$ moves closer to the requested target $z^\ast$ on the constrained axes $S$.
$R_{\text{hon}}$ rewards calibration: the model should self-report $\hat z_S \approx z_S(x)$.
Details are in App.~\ref{app:rl-config}.

\paragraph{Training.}
We fine-tune Qwen3-1.7B thinking~\citep{yang_qwen3_2025} with QLoRA adapters~\citep{hu_lora_2021,dettmers_qlora_2023} using group-relative policy optimization (GRPO) with GSPO-style sequence-level corrections~\citep{shao_deepseekmath_2024,deepseek-ai_deepseek-r1_2025,zheng_group_2025}.
Training follows a curriculum over constrained-axis subsets $S$ and target magnitudes (App.~\ref{app:curriculum}).
For code, we interleave two short warm-start (prediction) phases on the fixed $N{=}188$ program library (coordinate prediction and axis-level hypotheses) with full target-hitting training (App.~\ref{app:prompt-format}, App.~\ref{app:curriculum}).
Full fine-tuning and training hyperparameters are given in App.~\ref{app:rl-config}.

\paragraph{Story prompt sets.}
We train on 14 hand-written prompt templates and evaluate multi-branch diversity on 20 WritingPrompts prompts~\citep{fan_hierarchical_2018}. The training and evaluation prompt sets are disjoint.

\subsection{Use: target sweeps and outer-loop search in \texorpdfstring{$\mathbf{Z}$}{Z}}
\label{subsec:code-latent-design}

Once trained, the controller defines a black-box map $\mathcal{F}:(p,z^\ast)\mapsto x$ that supports both target sweeps and (when an evaluator is available) outer-loop optimization in $Z$. For stories, we sweep targets $z^\ast$ in $Z_{\text{text}}$ for a fixed prompt to obtain conditionally independent branches $x\sim\pi_\theta(\cdot\mid p,z^\ast)$ that can be generated in parallel without prompt chaining (Sec.~\ref{subsec:path-vs-state-diversity}).

In the code experiments, we instantiate the outer loop as BO with a Gaussian-process surrogate and expected improvement (GP--EI).

In the code domain, each completion implements a constrained \texttt{make\_seed() -> list[str]} and is executed and scored by the withheld CA++ benchmark $f(x)\in[0,1]$ (App.~\ref{app:ca-benchmark}).
The controller is objective-agnostic because $f$ is never used as an RL reward.
Instead, an outer loop proposes $z^\ast$, samples $K$ candidates, selects the best valid
program $x_{\text{best}}$, and (e.g.\ under BO) updates at the realised coordinate
$(z(x_{\text{best}}), f(x_{\text{best}}))$ rather than at the request $z^\ast$
(Alg.~\ref{alg:os-search}, App.~\ref{app:code-search}), since $z^\ast$ is a request and
reachable support in $Z_{\text{code}}$ is non-uniform (App.~Fig.~\ref{fig:code_zplane_score}).

\section{Results}\label{sec:results}

We first report target-tracking diagnostics in the frozen space $Z$ (Sec.~\ref{subsec:target-tracking}).
We then evaluate \emph{OS-Search} in two settings.
For stories, we treat the anchored $Z_{\text{text}}$ as a controllability and diversity interface (Sec.~\ref{subsec:z1-slop}, Sec.~\ref{subsec:path-vs-state-diversity}).
For code, we treat $Z_{\text{code}}$ as a search interface under the withheld CA++ objective (Sec.~\ref{sec:code-results}).

\subsection{Target-tracking accuracy in \texorpdfstring{$\mathbf{Z}$}{Z}}
\label{subsec:target-tracking}

Figure~\ref{fig:success_tolerance} shows that best-of-5 (Bo5) improves precision. Because the prompt schema is unchanged across checkpoints, the gap between early and late curves largely reflects learning from RL. In the code domain, the exemplar library also grows during RL, so gains reflect both a stronger policy and denser retrieval grounding. Code prompt ablations (App.~Tab.~\ref{tab:code_exemplar_ablation_eps005})
highlight the role of retrieval grounding. Removing exemplars or providing mismatched exemplars
collapses 3D control (Bo5 $\le 0.006$), and dropping exemplars also sharply reduces the fraction
of requests that yield any valid sample (ValidReq $=35.3\%$). By contrast, omitting the explicit
\texttt{REQUESTED TARGET} line has little effect (Bo1 $0.658$ vs.\ $0.664$ with the default prompt), which suggests that in code most conditioning is carried by exemplar selection rather than direct numeric-coordinate parsing. Calibration plots, with targets spanning $q_{0.01}$ to above $q_{0.99}$, show code tracking within $\pm0.05$ while stories are looser (Fig.~\ref{fig:calibration}, App.~\ref{app:accuracy}).

\begin{table*}[thb]
	\centering
	\small
	\resizebox{\textwidth}{!}{%
		\begin{tabular}{lcccccc}
			\toprule
			Method & Self-BLEU$\downarrow$ & ROUGE-L$\downarrow$ & METEOR$\downarrow$ &
			Sent-Sim$\downarrow$ & LLMScore$\uparrow$ & Degen$\downarrow$ \\
			\midrule
			Chained greedy ($T=0$) & 78.90 & 83.26 & 85.41 & 92.36 & 13.00 & \textbf{0.06} \\
			Chained temp sweep (best: $T=0.9,\,p=0.95,\,k=20$) & 58.45 & 66.20 & 70.31 & 87.89 & 17.35 & \textbf{0.06} \\
			Chained top-$p$ sweep (best: $T=0.6,\,p=0.95,\,k=20$) & 44.74 & 54.37 & 59.83 & 84.25 & 18.45 & 0.07 \\
			Chained top-$k$ sweep (best: $T=0.6,\,p=0.95,\,k=40$) & 63.96 & 71.05 & 74.82 & 89.38 & 13.95 & \textbf{0.06} \\
			\midrule
			\textbf{OS-Search ($Z$-grid sweep, $2\times3\times3$)} & \textbf{8.01} & \textbf{23.27} & \textbf{28.40} & \textbf{48.17} & \textbf{57.15} & 0.09 \\
			\bottomrule
		\end{tabular}
	}
	\caption[Path-based decoding vs sweeping targets in Z.]{\textbf{Path-based decoding vs sweeping targets in $Z$.}
		Multi-branch diversity metrics computed per prompt and then averaged across prompts.
		Path-based baselines generate $B=15$ branches per prompt via chaining with explicit negative examples.
		\emph{OS-Search} sweeps a fixed target grid in $Z_{\text{text}}$. For comparability we report $B=15$ branches per prompt (see App.~\ref{app:story-decoding-sweeps}).
		Lower scores on Self-BLEU~\citep{papineni_bleu_2002,zhu_texygen_2018}, ROUGE-L~\citep{lin_rouge_2004}, METEOR~\citep{banerjee_meteor_2005}, and Sent-Sim indicate higher
		diversity, while higher LLMScore and lower Degen are desirable.}
	\label{tab:z-vs-sampling-diversity}
\end{table*}

\subsection{Anchored axis \texorpdfstring{$\mathbf{z_1}$}{z1} and EQ-Bench Slop-Score}
\label{subsec:z1-slop}

We use EQ-Bench Slop-Score~\citep{paech_eq-bench_2024,paech_slop-score_2025} as a proxy for templated lexical patterns (colloquially ``AI slop''). Because $z_1$ is explicitly anchored toward Qwen3-generated ``default-style'' stories relative to the corpus mean, we use Slop-Score primarily as a post-hoc axis-validation diagnostic (not as a general text-quality metric). App.~\ref{app:z1-diagnostics} summarises analyzer details, and provides qualitative excerpts spanning realised $z_1$.

\begin{figure}[htb]
	\centering
	\includegraphics[width=\linewidth]{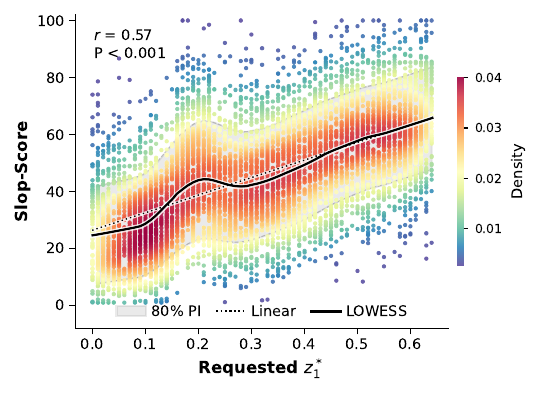}
	\caption[Relationship between requested z1* and Slop-Score.]{\textbf{Relationship between requested $z_1^\ast$ and Slop-Score.}
		Slop-Score as a function of the requested target on the first axis,
		$z_1^\ast$, for completions from late-training rollouts (final 10\% of GRPO updates) generated under	1D control ($S=\{1\}$) ($N \approx 7{,}000$). Points are coloured by local density.
		The solid line shows a LOWESS smooth, the dotted line the global linear
		regression fit ($r \approx 0.57$), and the shaded band an
		approximate 80\% prediction interval around the LOWESS trend.}
	\label{fig:fig_z1_slop_main}
\end{figure}

For final-10\% 1D-control rollouts ($S=\{1\}$, $N \approx 7{,}000$), Slop-Score correlates with the requested target $z_1^\ast$ with Pearson $r \approx 0.57$ ($p < 10^{-3}$). Fig.~\ref{fig:fig_z1_slop_main} shows an approximately linear trend: increasing $z_1^\ast$ shifts the Slop-Score distribution upward by roughly 20--30 points over our target range, with substantial within-setting variability.
To test whether this association is specific to the anchored axis, we also correlate Slop-Score with the realised coordinates $(z_1,z_2,z_3)$ on the same samples (App.~Fig.~\ref{fig:slop_scores_realized}). Only $z_1$ shows a substantial association ($r \approx 0.56$), while correlations with $z_2$ and $z_3$ are small. We do not explore negative $z_1$ targets. In our STORIES corpus this half-space is dominated by low-structure diary-style prose, and pushing the small Qwen3-1.7B backbone into $z_1 \leq 0$ reduces coherence.

\subsection{Story diversity under path-based baselines versus state-like target sweeps}
\label{subsec:path-vs-state-diversity}

We compare state-like target sweeping to a strong path-based multi-branch baseline on WritingPrompts~\citep{fan_hierarchical_2018}. As context, Avoidance Decoding applies a decoding-time similarity penalty and reports strong gains under its own harness~\citep{park_avoidance_2025} (see Sec.~\ref{sec:related_work}). Because our backbone, prompt subset, and LLM judge differ, we restrict quantitative comparisons to our fixed harness.

\paragraph{Baselines.}
We compare to a negative-example prompt-chaining baseline used by \citet{park_avoidance_2025} on the base Qwen3-1.7B thinking model with decoding-parameter sweeps to obtain $B=15$ branches per prompt. Details are in App.~\ref{app:story-decoding-sweeps}.

\paragraph{OS-Search.}
We generate branches with the \emph{OS-Search} controller $\pi_\theta(\cdot\mid p,z^\ast)$ using a fixed sampler and varying only the requested target $z^\ast$ in $Z_{\text{text}}$ (details in App.~\ref{app:story-decoding-sweeps}). Story targeting is coarse at tight tolerances (Fig.~\ref{fig:success_tolerance}), but our grid sweep uses well-separated target settings. This is sufficient to push different branches into distinct regions of the frozen space and drives the diversity gains in Tab.~\ref{tab:z-vs-sampling-diversity}.

Tab.~\ref{tab:z-vs-sampling-diversity} shows that sweeping targets in $Z_{\text{text}}$ yields substantially higher diversity while keeping degeneration low: the strongest sampling baseline achieves $\text{LLMScore}=18.45$ with $\text{Degen}=0.07$, while \emph{OS-Search} achieves $\text{LLMScore}=57.15$ with $\text{Degen}=0.09$. These gains are consistent across lexical and embedding overlap metrics, indicating substantially greater lexical and embedding-level diversity. Using the ratio $\rho_{\text{div}}$ defined in App.~\ref{app:story-decoding-sweeps}, we obtain
\[
\rho_{\text{div}}(\text{OS-Search})
\;=\;
\frac{57.15}{18.45}
\approx 3.10.
\]

\subsection{Code-space search in \texorpdfstring{$Z_{\text{code}}$}{Z code} under a withheld CA++ objective}
\label{sec:code-results}

\begin{figure}[bt]
	\centering
	\includegraphics[width=\linewidth]{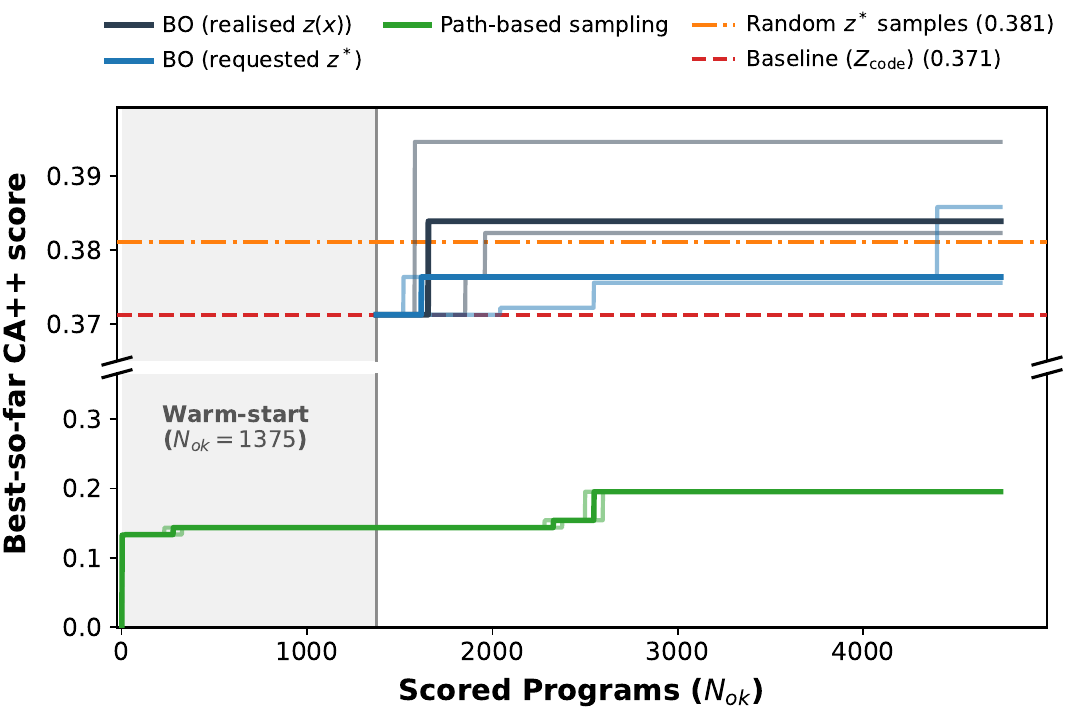}
	\caption[Best-so-far CA++ score under OS-Search (matched budget).]{\textbf{Best-so-far CA++ score vs scored-program budget.}
		$N_{\mathrm{ok}}$ counts scored valid programs (invalid/malformed outputs are discarded).
		The grey segment indicates the warm-start phase for BO ($N_{\mathrm{ok}}=1375$).
		Thick lines show the median over three seeds, and thin lines show the min/max range (split y-axis for readability).}
	\label{fig:code_bo_trace}
\end{figure}

We evaluate the code-domain \emph{OS-Search} interface described in Sec.~\ref{subsec:code-latent-design} on the withheld CA++ benchmark $f(x)\in[0,1]$ (App.~\ref{app:ca-benchmark}).
\begin{figure}[h]
	\centering
	\includegraphics[width=\linewidth]{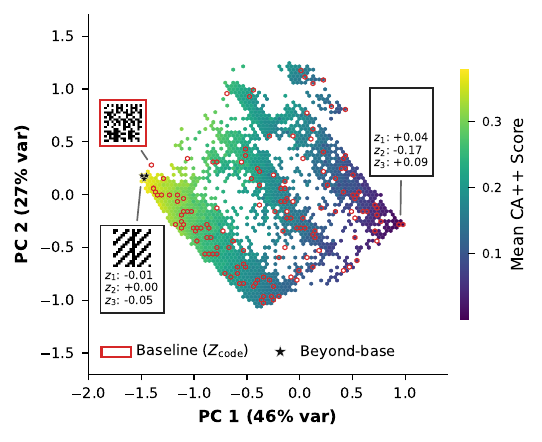}
	\caption[Programs in CA++ behavioral space (code domain).]{\textbf{Programs in CA++ behavioral space (code domain).}
		Each point is a valid \texttt{make\_seed()} program scored by CA++.
		Axes are PCs of the CA++ trajectory-statistic feature vectors of the plotted programs (so proximity indicates similar CA dynamics), not $Z_{\text{code}}$.
		Hexbins show random-$z^\ast$ controller samples coloured by mean score per bin. 
		Red outlines mark bins reached by the $N{=}188$ program-library baseline, and
		stars mark bins whose mean exceeds the best baseline score (0.371). Insets show representative seeds generated by corresponding \texttt{make\_seed()} implementations. The printed $(z_1,z_2,z_3)$ values are the requested targets $z^\ast$ for those programs.}
	\label{fig:code_ca_hexbin}
\end{figure}
\paragraph{Budget accounting and baselines.}
We measure evaluation budget by the number of scored valid programs, denoted $N_{\mathrm{ok}}$.
This counts candidates that parse, execute, return a valid board, and receive a CA++ score.
Our only budget-matched baseline is path-based sampling from the base model
(no $z^\ast$, Fig.~\ref{fig:code_bo_trace}, green).
For reference (i.e., not budget-matched to the online search runs below), we also report the fixed $N{=}188$ program-library sweep used to fit $Z_{\text{code}}$ (best $0.371$),
and a larger precomputed pool of random-$z^\ast$ controller rollouts (valid $N_{\mathrm{ok}}{=}13{,}752$, best $0.381$), which we use to contextualize coverage
and attainable scores.

\paragraph{Coverage and beyond-baseline regions.}
Random-$z^\ast$ sampling reaches a broader region of CA behavioral space and uncovers compact high-score pockets beyond the library frontier (Fig.~\ref{fig:code_ca_hexbin}).

\paragraph{Outer-loop optimization over $z^\ast$: surrogate-update ablation.}
We run warm-started BO (GP--EI) over requested targets $z^\ast$ (App.~\ref{app:code-search}). Consistent with the realised-coordinate interface described in Sec.~\ref{subsec:code-latent-design}, updating the surrogate at the realised coordinates $z(x_{\text{best}})$ of the best valid program per query outperforms updating at the request $z^\ast$ in our surrogate-update ablation (App.~Tab.~\ref{tab:bo_ablation}). Under a matched $N_{\mathrm{ok}}$ budget, BO improves the best-so-far CA++ score up to $0.395$ (Fig.~\ref{fig:code_bo_trace}), compared to $0.371$ for the library and well above path-based sampling. For reference, Fig.~\ref{fig:bo_best_seed_stack} visualizes the best scoring seed ($0.395$). Additional Life-torus diagnostics are in App.~\ref{app:code-search}.

\section{Discussion}

\emph{OS-Search} makes $z^\ast$ a low-dimensional control knob: after training a controller to approximately hit requested coordinates in a frozen space $Z$, we can explore or optimize outputs by searching over targets in $Z$ while keeping inference as standard autoregressive sampling.

Empirically, the interface supports two regimes. For stories, a small grid sweep in the anchored $Z_{\text{text}}$ produces diverse, non-degenerate branches without prompt chaining (Tab.~\ref{tab:z-vs-sampling-diversity}), and the anchored axis provides a stable reference axis (Fig.~\ref{fig:fig_z1_slop_main}). For code, treating $Z_{\text{code}}$ as a design space enables black-box optimization: random target sampling expands coverage and GP--EI over $z^\ast$ improves a withheld executable objective (Fig.~\ref{fig:code_bo_trace}). These results suggest that outer-loop search should model actuator noise and update on realised $z(x)$ rather than on requests $z^\ast$ (App.~Tab.~\ref{tab:bo_ablation}).

More broadly, \emph{OS-Search} is a modular ``outer-loop + evaluator'' interface that is complementary to decoding-time control: it shifts work to training so inference can remain standard sampling, and it is most useful when a meaningful space and validity/evaluation signals exist. We expect larger backbones and richer exemplar sets may improve prompt-only use of targets. Our core contribution is the fixed encoder-defined output space $Z$ as a per-sample search/control interface, with GRPO/GSPO training serving as one practical way to make $z^\ast$ reliably actionable. Next steps include better spaces, stronger actuators, and outer loops beyond Bayesian optimization. 

Because group-based RL objectives such as GRPO/GSPO depend on within-prompt diversity for informative relative rewards, \emph{OS-Search} may also be useful during RL by replacing ``turn up temperature'' with structured target sweeps that deliberately span distinct solution or reasoning modes (e.g., sampling around an initial chain-of-thought-trajectory's $z(x)$ or running a small per-prompt BO on extremely hard prompts).

\section{Limitations}
\label{sec:limitations}

\emph{OS-Search} is an interface and feasibility study, and it comes with important limitations.

\paragraph{Code release scope.}
Code to construct the frozen output spaces and reproduce the code-domain experiments (including CA++ and the BO loop) is available.
We intentionally do not release the story-domain training pipeline, datasets, or checkpoints due to the text-domain safety and dual-use risks discussed below.
The released repository is not a mirror of the internal research codebase; components were removed and interfaces simplified for release, so story-domain scripts referenced in the paper are not included.

\paragraph{Extreme tail targeting in text can elicit unsafe content.}
In text domains, the learned control axes in a frozen encoder-defined space may align with broad affective or genre-related factors in the underlying corpus. We observed that pushing requests into extreme tail regions (e.g., large-magnitude $z^\ast$ on non-anchored axes such as $z_2$/$z_3$) can occasionally produce highly negative or graphic content. This can be amplified by retrieval grounding, since nearest-neighbour exemplars in those regions may themselves exhibit the same style and provide in-context steering toward it. Our training objective rewards target tracking and calibration but does not include an explicit content-safety term, so such behaviours are not disincentivized by default.

\paragraph{Risks of extreme text targeting and dual use.}
OS-Search exposes a low-dimensional, reusable coordinate interface $Z$ together with an outer-loop search mechanism (sweeps or black-box optimization over $z^\ast$). In text domains, this interface can make it easier to systematically search for generations with particular stylistic or rhetorical properties than ad-hoc prompt engineering, because the outer loop can iterate over targets without growing context and can reuse the same actuator across prompts. In the wrong hands, such search could be used to optimize for undesirable behaviors when paired with an automated evaluator. While our story experiments focus on diversity and on validating an anchored axis using Slop-Score as a diagnostic (not as an optimization target), richer spaces, larger backbones, or different anchor choices could enable more extreme forms of style/persona targeting. Practical deployments should therefore incorporate strong content filtering, careful curation/guardrails on the exemplar library used for retrieval grounding, monitoring for adversarial outer-loop objectives, and conservative bounds on target search regions. These risks are not unique to OS-Search, but the state-like target interface can reduce the friction of iterative search in generation space.

\paragraph{Meaning of $Z$ depends on the encoder and corpus.}
The geometry and semantics of $Z$ depend on the frozen encoder $E$, the corpus used to fit
$(\mu,U)$, and embedding/preprocessing choices (e.g., chunking and pooling). Changing any
of these can change neighborhoods, axis structure, and which regions are reachable or
useful. This dependence is the price of having a stable external interface: portability to new domains requires rebuilding and revalidating the space.

\paragraph{Anchoring is pragmatic, not canonical.}
In $Z_{\text{text}}$ we anchor $z_1$ with a small set of ``default-style'' stories to
obtain a knob with a clear empirical correlate. Different anchors would produce different
axes and may encode different stylistic biases. Beyond the anchored axis, PCA+Varimax can
encourage simpler directions but does not guarantee human-interpretable factors.

\paragraph{Control is approximate and reachability is non-uniform.}
A requested target $z^\ast$ is a request. Realised coordinates occupy a smaller,
non-uniform subset of $Z$, and some target combinations may be effectively unattainable or
attained only with low probability (Sec.~\ref{subsec:code-latent-design}).
This is most visible in stories, where strict 3D targeting is coarse at tight tolerances
(Fig.~\ref{fig:success_tolerance}). Best-of-$K$ sampling can improve targeting, but it
increases inference cost. For optimization, the outer loop should model actuator noise and
reachable support rather than assuming perfect control.

\paragraph{Dependence on retrieval-grounded prompting.}
Targets are grounded by retrieving exemplars near $z^\ast$ (and a contrast example near $-z^\ast$) under the same frozen space (Fig.~\ref{fig:prompt-schema}). Performance therefore depends on library coverage and on the prompt format’s interaction with the base model. Ablating exemplar retrieval substantially degrades targeting and (in code) validity (App.~Tab.~\ref{tab:code_exemplar_ablation_eps005}). A more exhaustive ablation would further isolate retrieval-only effects and quantify sensitivity to exemplar count, selection strategy, and library quality.

\paragraph{Text evaluation focuses on probes and diversity, not end-to-end optimization.}
We use Slop-Score mainly as a axis-validation diagnostic (Sec.~\ref{subsec:z1-slop}) and evaluate branching with overlap metrics and an LLM judge (Tab.~\ref{tab:z-vs-sampling-diversity}).
These metrics are informative but are not task-level objectives, and LLM-based judging can
introduce model-dependent bias. We do not demonstrate an end-to-end outer-loop text task analogous to the code-domain optimization.

\paragraph{Baselines and domains are not exhaustive.}
Our story baselines are strong within our fixed harness (prompt-chaining with explicit
negative examples plus decoding sweeps), but they do not cover the full space of decoding-time
diversity methods or alternative branching pipelines. Broader comparisons would sharpen the
empirical positioning.

\paragraph{The code benchmark is synthetic and heavily constrained.}
CA++ provides a clean executable testbed, but it is not representative of many real
software engineering settings. Our programs are single-function and tightly constrained,
and the validity gate is correspondingly specific. Extending \emph{OS-Search} to richer objectives raises additional challenges in validity checking, space construction, and search efficiency.
We chose this setting after piloting several code objectives with the 1.7B parameter backbone and a limited GPU-memory budget: generating valid $16{\times}16$ boards reliably stayed within the model's capability and kept completion lengths manageable, whereas more realistic tasks often collapsed to very low validity rates and/or produced long \xtag{think} traces that were infeasible to train with under our constraints.

\paragraph{Compute, scaling, and safety considerations.}
\emph{OS-Search} shifts effort to training (sequence-level RL plus repeated encoder calls) and
outer-loop optimization may require many valid, scored samples when validity rates are low.
Any method that enables systematic exploration of an LLM’s output space can also be misused
to search for undesirable behaviours if paired with an inappropriate evaluator. Practical
deployments should combine this interface with domain-appropriate safety filters,
constraints, and monitoring.

\section*{Resources}
Code (partial release: $Z$-space construction + code-domain training): \url{https://github.com/TobiasMaterzok/OS-Search}. 
Story-domain training code/data are intentionally not released; see \S\ref{sec:limitations}.

{\raggedright

}

\appendix

\section*{\huge Appendices}

\section{Output-space construction and evaluation}
\label{app:zspace-details}

We detail the construction of the frozen 3D output coordinates used throughout the paper.
For each domain we fix a text/code encoder $E(\cdot)$, fit a linear projection, and define
\[
z(x) \;=\; U^\top\!\big(E(x)-\mu\big)\in\mathbb{R}^{d_z}, \qquad d_z=3,
\]
with corpus mean $\mu$ and an orthonormal basis $U\in\mathbb{R}^{D\times d_z}$. We freeze $(E,\mu,U)$ for all RL experiments.

\paragraph{Corpora.}
\textbf{$Z_{\text{text}}$ (stories).}
We combine multiple public English story datasets (short stories and longer narratives), including the STORIES corpus of \citet{trinh_simple_2019}. We keep only text fields and filter documents to roughly $800$--$25{,}000$ characters.

\textbf{$Z_{\text{code}}$ (code).}
We build a task-specific library of $N=188$ diverse Python implementations of the constrained function \texttt{make\_seed() -> list[str]} and fit $Z_{\text{code}}$ on this set.
We use a task-specific corpus because spaces fit on general-purpose code corpora tended to allocate axes to variations that are irrelevant or disallowed under our \texttt{make\_seed} constraints.

\paragraph{Encoders and pooling.}
Stories are embedded with a \texttt{sentence-transformers} encoder based on all-mpnet-base-v2~\citep{reimers_sentence-bert_2019,song_mpnet_2020}. Code is embedded with jina-embeddings-v2-base-code~\citep{gunther_jina_2024}.
For long inputs we use automatic chunking. We mean-pool chunk embeddings and apply $\ell_2$ normalization.

\paragraph{PCA and rotation template (both domains).}
Given a corpus of $N$ texts/programs, we embed to a matrix $X\in\mathbb{R}^{N\times D}$ and set $\mu=\frac{1}{N}\sum_i X_i$.
We run PCA on the centered embeddings to an intermediate dimension $d_{\text{inter}}$, yielding orthonormal components
$C_{\text{inter}}\in\mathbb{R}^{D\times d_{\text{inter}}}$ and eigenvalues $\lambda_1,\dots,\lambda_{d_{\text{inter}}}$.
Optionally, we whiten inside the PCA space with
$s_j=(\lambda_j+\varepsilon)^{-1/2}$ and $S=\mathrm{diag}(s_1,\dots,s_{d_{\text{inter}}})$.
We then choose $d_z=3$ directions inside this intermediate space (with anchoring for stories, no anchoring for code), map them back to the encoder space, and orthonormalize to obtain $U$.

Finally, we apply an orthogonal rotation within the retained $d_z$-dimensional subspace (Varimax, anchored Varimax for stories) to encourage simpler axis loadings. This preserves the projector $UU^\top$ and changes only the reported axis orientation. For code, we apply this template with no anchoring, fitting on the $N=188$ valid \texttt{make\_seed()} library.

\paragraph{Anchored story space $Z_{\text{text}}$.}
To make one axis have a stable semantic reference, we anchor the first axis toward a small set of Qwen3-generated ``default-style'' short stories.
Let $\mathcal{A}$ be the anchor set and $a=\frac{1}{|\mathcal{A}|}\sum_{x\in\mathcal{A}}E(x)$ be their mean embedding. Define the anchor direction relative to the corpus mean as $a_{\mathrm{dir}}=a-\mu$.
We project this direction into the intermediate PCA space (and apply whitening if enabled) and normalize it to obtain a unit anchor $w\in\mathbb{R}^{d_{\text{inter}}}$.
We then choose the remaining $(d_z-1)$ axes to capture as much residual PCA variance as possible subject to orthogonality to $w$ (i.e., we take the leading eigenvectors of the PCA-variance matrix after projection onto $w^\perp$).
After mapping back and orthonormalization, we apply an anchored Varimax rotation: we keep the first axis aligned with the anchor direction and Varimax-rotate only the remaining $d_z-1$ axes so that $z_2$ and $z_3$ are easier to interpret while preserving the anchored $z_1$.

\paragraph{Model selection and diagnostics.}
We select hyperparameters via a small grid search over:
$d_{\text{inter}} \in \{8,10,12,14,16,18,20,24,28,32,36,40,48\}$,
$d_z \in \{3,4,5,6,7,8\}$ (with $d_z \le d_{\text{inter}}$),
whitening on/off, and (for stories) anchoring on/off.
For each configuration we compute:
\begin{itemize}
	\item \textbf{Reconstruction:} mean $\|x-\hat{x}\|_2$ where $\hat{x}=\mu+UU^\top(x-\mu)$, and reconstruction $R^2$ (relative to centered embeddings).
	\item \textbf{Neighborhood preservation:} $k$-NN recall and trustworthiness in $Z$ at $k\in\{10,20\}$.
	\item \textbf{Axis diagnostics:} per-axis variance and mean Hoyer sparsity (averaged across axes)~\citep{hoyer_non-negative_2004}.
	\item \textbf{Stories only:} anchor capture $\|U^\top a_{\mathrm{dir}}\|_2/\|a_{\mathrm{dir}}\|_2$ and the fraction of captured anchor energy that lies on $z_1$.
\end{itemize}
A scalar preference score combines these quantities with fixed weights (reported in the code release) and a mild quadratic penalty on $(d_z-3)^2$. For anchored story configurations, we discard settings whose anchor capture is below a fixed threshold or whose captured anchor does not load primarily onto $z_1$. The best configuration per domain is used in all subsequent experiments.

\section{Prompt format, parsing, and embedding pipeline}
\label{app:prompt-format}

\paragraph{Structured envelope and parsing.}
Completions follow the structured envelope in Fig.~\ref{fig:prompt-schema}.
A lightweight parser enforces the required fields and checks that \xtag{target} contains numeric values for the requested axes.
Malformed outputs receive zero reward via $R_{\text{format}}$.
For all downstream computations we ignore \xtag{think} and \xtag{title} and embed only the task-content field \xtag{text}.

\paragraph{Retrieval grounding for numeric targets.}
Given a base prompt $p$ and a requested target $z^\ast_S$, we retrieve exemplars from the corresponding library indexed in the frozen space using Euclidean nearest neighbours in $Z$:
two exemplars nearest to $z^\ast$ and one contrast exemplar nearest to $-z^\ast$.
We insert these exemplars together with the numeric request into the prompt (Fig.~\ref{fig:prompt-schema}, Sec.~\ref{subsec:zspace-rl}).
In exemplar blocks, \xtag{target} contains stored realised coordinates $z(x_j)$, in model outputs, \xtag{target} contains the self-report $\hat z_S$.

\paragraph{Computing realised coordinates $z(x)$.}
After generation, realised coordinates are computed by embedding only \xtag{text} and projecting into the corresponding frozen space.
For stories, we embed \xtag{text} with $E_{\text{text}}$ and project to $Z_{\text{text}}$.
For code, we deterministically extract the \texttt{make\_seed()} function definition from \xtag{text}, strip comments, embed the extracted code with $E_{\text{code}}$, and project to $Z_{\text{code}}$.

\paragraph{Code constraints, validation, and scoring.}
In the code domain, \xtag{text} must implement a single function \texttt{make\_seed() -> list[str]} that returns exactly 16 strings of length 16 over \texttt{'.'} and \texttt{'\#'}.
We extract \texttt{make\_seed()} with a deterministic parser and treat a completion as valid only if it passes AST sanitization and executes in a restricted environment to return a correctly formatted $16\times16$ board.
Invalid candidates are discarded (and receive zero reward).
Valid programs are scored by the cellular-automaton evaluator in App.~\ref{app:ca-benchmark}.

In two brief warm-start stages, we train on library programs by asking the model to state brief hypotheses about which code features influence each axis and to predict the program's $z$-coordinates, using far-away library examples as context to prevent trivial copying.

\paragraph{Curriculum and reward pointers.}
Across both domains, training requests vary the constrained-axis subset $S$ and target values $z^\ast_S$ according to App.~\ref{app:curriculum}.
Reward components (format checks, distance shaping, and self-report scoring) are implemented as detailed in App.~\ref{app:rl-config}.

\section{Curriculum and target sampling}
\label{app:curriculum}

\paragraph{One training ``request'' and phase budget.}
We define a request as a sampled pair $(S, z^\ast)$. The set $S \subseteq \{1,2,3\}$ specifies which axes are constrained, and $z^\ast$ gives the requested target in the frozen space. Because the controller is trained with the retrieval-grounded prompt schema, we can request arbitrary $z^\ast$ values and retrieve the closest exemplars from the current library to instantiate the prompt.

Each target-hitting curriculum phase contains $15{,}000$ distinct prompt--target requests.
Concretely, we run $1{,}500$ GRPO optimizer steps with gradient accumulation $10$.

\paragraph{Phase schedule (stories vs.\ code).}
The curriculum differs slightly by domain:
\begin{itemize}
	\item \textbf{Stories:} three phases that increase control dimensionality:
	$S=\{1\}$, then $S=\{1,2\}$, then $S=\{1,2,3\}$.
	\item \textbf{Code:} an interleaved procedure:
	(i) a warm-start phase on the fixed $N=188$ \texttt{make\_seed} library to train
	coordinate prediction and axis-level hypotheses, then
	(ii) target-hitting training with full 3D control $S=\{1,2,3\}$, then
	(iii) a second brief warm-start phase (same prediction format), then
	(iv) a final target-hitting phase.
\end{itemize}

\paragraph{Exemplar library initialization and growth.}
We initialize the exemplar library by embedding a fixed corpus and keeping
documents from the outer quantile bands of each axis, so the initial library
covers mostly extreme regions of $Z$.
For code, we additionally grow this library during GRPO by adding valid
model-generated \texttt{make\_seed} programs. During RL we also penalize
near-duplicate copying of retrieved exemplars (App.~\ref{app:rl-config}).

\begin{figure*}[htb]
	\centering
	\includegraphics[width=\linewidth]{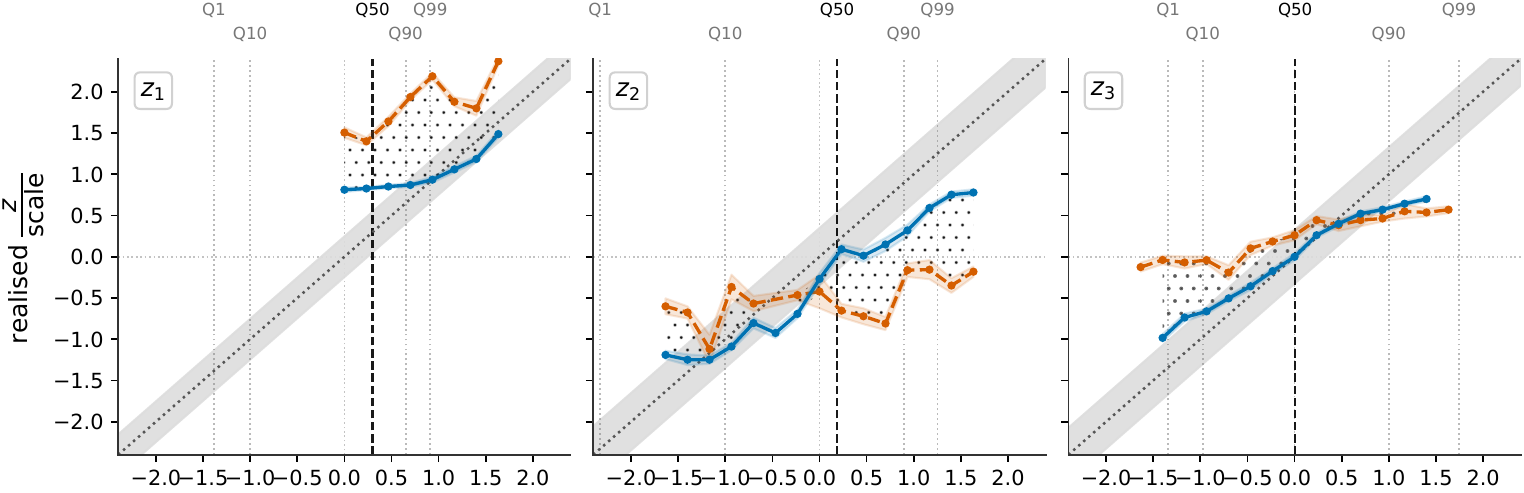}\\[+0.15em]
	\includegraphics[width=\linewidth]{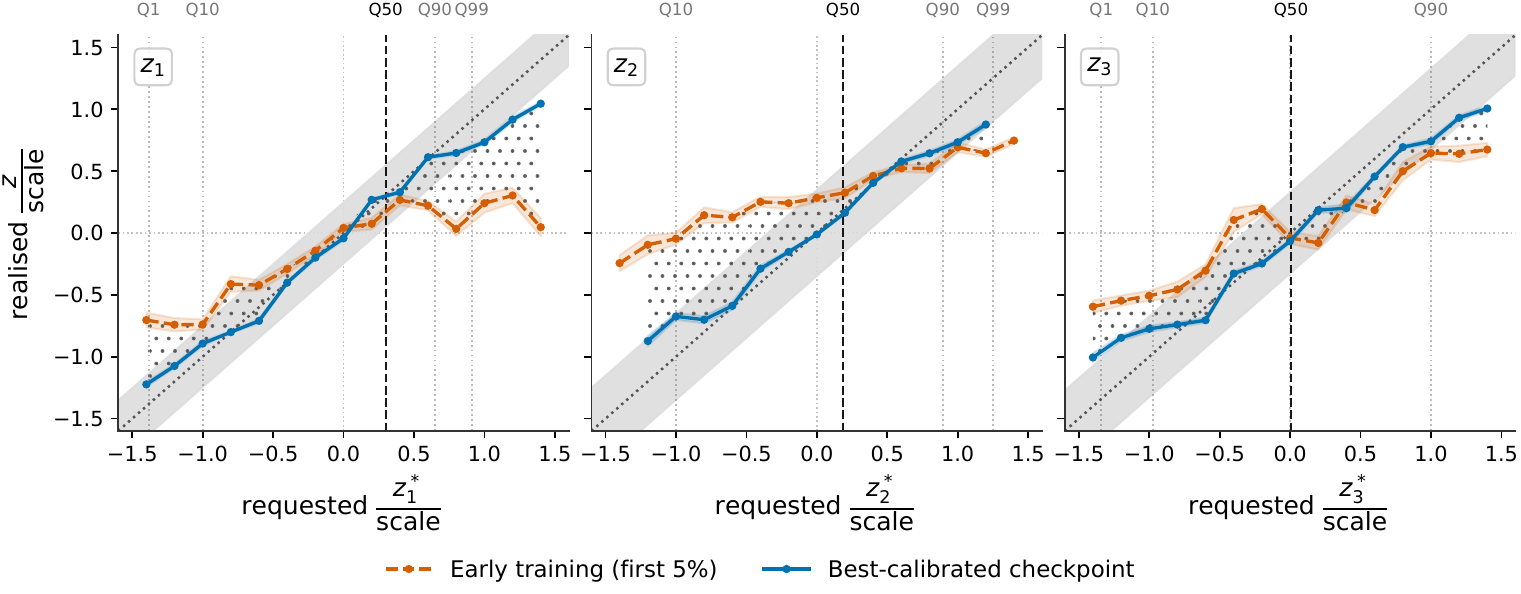}
	\caption[Calibration of state-like control in Z.]{\textbf{Calibration of state-like control in $Z$.}
		Binned mean realised coordinate $z_i$ versus requested target $z_i^\ast$ for each constrained axis.
		The diagonal indicates perfect calibration.
		Blue: best checkpoint. Orange: early/baseline (first 5\% of RL).
		The grey band marks the reference tolerance $|z_i-z_i^\ast|\le 0.05$ (raw $z$ units).
		Vertical markers show reference-corpus quantiles $(q_{0.01},\dots,q_{0.99})$ of $z_i$
		Curves are computed from $7{,}000$ sampled evaluation completions per domain
		(usable records: stories $N=6{,}946$, code $N=6{,}860$).
		\textbf{Top:} story policy, fixed exemplar library. \textbf{Bottom:} code policy, exemplar library grows during RL.}
	\label{fig:calibration}
\end{figure*}

\paragraph{Frozen axis statistics and target ranges.}
We precompute frozen per-axis statistics from a large embedded reference set (empirical quantiles and IQR). For each axis $i$ we define a robust scale
\[
s_i = \max\big(|q_{0.10}|,\;|q_{0.90}|,\;10^{-4}\big),
\]
and freeze $(s_1,s_2,s_3)$ for the entire training run. These scales are used both to sample requested targets and to normalize distance violations in the reward. Targets are drawn from an axis-aligned hypercube whose side lengths are proportional to $(s_i)$ by choosing a sign and sampling a magnitude from a fixed band of multiples of $s_i$, intentionally reaching into tail regions (often up to and beyond $q_{0.99}$ on skewed axes). For context, Fig.~\ref{fig:calibration} overlays reference-corpus quantiles $(q_{0.01},\dots,q_{0.99})$. We request magnitudes up to $1.75\,s_i$ for text and $1.5\,s_i$ for code, so some requests lie beyond $q_{0.99}$ on heavy-skew axes, i.e., we deliberately train on extrapolation into rare tail regions of the corpus.

\paragraph{Prototype targets and within-phase shaping.}
Early in training we anchor sampling on a small set of prototype centers in $Z$
(a few dozen sign patterns over the axes, scaled by the frozen $(s_i)$).
Later batches draw targets by sampling magnitudes from a bounded band around the
origin and choosing a random subset of axes to constrain, with a bias toward
non-negative $z_1$ for stories.

\begin{figure}[h]
	\centering
	\includegraphics[width=\linewidth]{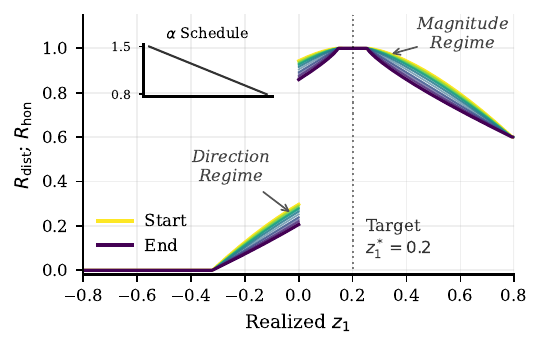}
	\caption[Evolution of the curriculum reward surface.]{\textbf{Evolution of the curriculum reward surface.}
		Reward profiles for a target value $z_1^\ast = 0.2$ are shown within a
		single curriculum phase. In each phase, the distance exponent
		$\alpha$ is linearly annealed from $1.5$ to $0.8$ over the
		1\,500 RL updates (inset), progressively narrowing the reward
		basin in the magnitude regime to demand higher precision later in
		that phase, while maintaining a distinct penalty step in the
		direction regime (sign mismatch). At the start of every new phase,
		$\alpha$ is reset to $1.5$.}
	\label{fig:fig_reward_surface_evolution}
\end{figure}

Each request also carries a per-example distance exponent $\alpha$
(Fig.~\ref{fig:fig_reward_surface_evolution}) that controls how sharply reward
concentrates around the target. Within each curriculum phase, $\alpha$ is
linearly annealed from $1.5$ to $0.8$ over the $1{,}500$ RL updates, then reset
to $1.5$ at the start of the next phase (from roughly MSE-like to more sharply
peaked near the target).

\section{RL configuration and reward implementation}
\label{app:rl-config}

\paragraph{Reward implementation.}
We compute $R_{\text{format}}$, $R_{\text{dist}}$, and $R_{\text{hon}}$ as defined in Sec.~\ref{subsec:zspace-rl}, using the parsing/embedding pipeline in App.~\ref{app:prompt-format} and frozen target scales from App.~\ref{app:curriculum}. Samples failing the format/validity gate receive reward zero.
Both $R_{\text{dist}}$ and $R_{\text{hon}}$ are down-weighted when the \xtag{think} trace or \xtag{text} block are very short, so trivial outputs receive no distance or honesty credit.
In addition, we scan \xtag{text} for explicit mentions of the control interface (e.g., mentions of $Z$ or axis coordinates).
If such a leak is detected we skip the embedding step and set $R_{\text{dist}} = R_{\text{hon}} = 0$.

For code, when exemplar retrieval is used, we additionally penalise copying of retrieved guidance examples. Concretely, we detect near-duplicate outputs relative to any retrieved exemplar and apply zero reward to discourage verbatim replication of the guidance library.

\paragraph{Sampling and training.}\label{app:unsloth-lora}
To enable fine-tuning under our computational constraints, we use Unsloth~\citep{han_unsloth_2023} with 4-bit Quantized Low-Rank Adaptation (QLoRA)~\citep{hu_lora_2021,dettmers_qlora_2023}.
We set the LoRA rank to $r = 64$ and scaling factor $\alpha = 64$, applying LoRA to all linear layers.

For RL, we use vLLM for efficient sampling~\citep{kwon_efficient_2023} with temperature $0.8$, top-$p = 0.95$,
and top-$k = 40$ for text. For code we use temperature $1.0$, top-$p = 0.95$, and top-$k = 50$.
We generate up to a fixed maximum number of tokens per completion (1600 token output limit) and stop at the model’s end-of-sequence token.

We train QLoRA adapters with per-device batch size $1$, $7$ generations per
prompt, $10$ gradient accumulations, learning rate in the range $2\cdot 10^{-5}$,
AdamW~\citep{kingma_adam_2017,loshchilov_decoupled_2019} with weight decay $0.01$, and a constant-with-warmup schedule $0.1$.
We use group-relative policy optimization (GRPO)~\citep{shao_deepseekmath_2024,deepseek-ai_deepseek-r1_2025} with GSPO-style sequence-level importance sampling and sequence-level clipping~\citep{zheng_group_2025}, using a clipping window $(\varepsilon,\varepsilon_{\text{high}}) = (0.01, 0.03)$.
Reward scaling is disabled following DR-GRPO~\citep{liu_understanding_2025}, so the geometry of the composite reward is preserved.

\section{Targeting accuracy and calibration}
\label{app:accuracy}

We report targeting accuracy for the \emph{OS-Search} controllers (text and code) and calibration diagnostics for the
frozen output spaces $Z_{\text{text}}$ and $Z_{\text{code}}$.

\paragraph{Metrics.}
For a sample $x\sim \pi_\theta(\cdot\mid p,z^\ast)$ with constrained-axis set
$S\subseteq\{1,2,3\}$, define the per-axis absolute error
$e_i(x,z^\ast)=|z_i(x)-z_i^\ast|$ for $i\in S$ and the joint (worst-axis) error
$e(x,z^\ast;S)=\max_{i\in S} e_i(x,z^\ast)$.
We use $\varepsilon=0.05$ as a tight reference tolerance (``$\pm0.05$ per-axis'' under the
$\ell_\infty$ joint error). All errors are in raw $z$ units.

For per-axis results we report $\mathrm{Success}@\varepsilon=\mathbb{P}[e_i\le \varepsilon]$
estimated over constrained axis instances.
For joint results we report $\mathrm{Success}@\varepsilon=\mathbb{P}[e\le \varepsilon]$
estimated over requests (with the relevant constrained set $S$).

\paragraph{Evaluation protocol.}
We evaluate each domain on $7{,}000$ sampled requests.
A completion is counted as usable if it passes the output-envelope parser and yields
a valid \xtag{text} block that can be embedded (and for code, the extracted
\texttt{make\_seed()} implementation must pass AST sanitization and execute to return a valid
$16\times16$ board). Calibration curves and error summaries below are computed on usable
completions (stories: $N=6{,}946$, code: $N=6{,}860$).

\paragraph{Calibration curves.}
Fig.~\ref{fig:calibration} reports binned mean realised coordinates $z_i(x)$ versus requested targets $z_i^\ast$ for each constrained axis, for an early checkpoint and the best checkpoint.
Because the prompt interface is unchanged across checkpoints (numeric targets plus exemplar
retrieval), the improvement reflects sequence-level RL training. In the code domain it also
reflects the growing exemplar library used for retrieval during RL.

\paragraph{Absolute targeting error at $\varepsilon=0.05$.}
The error summaries in Tabs.~\ref{tab:per_axis_accuracy_eps005} and \ref{tab:joint_accuracy_eps005} are computed on the same $7{,}000$ sampled evaluation requests used for the calibration curves in Fig.~\ref{fig:calibration}.
Tab.~\ref{tab:per_axis_accuracy_eps005} reports per-axis error statistics and per-axis
$\mathrm{Success}@0.05$ over constrained axis instances.
Tab.~\ref{tab:joint_accuracy_eps005} reports joint error under full 3D control
($S=\{1,2,3\}$), matching the joint-error definition used in Fig.~\ref{fig:success_tolerance}.

\begin{table}[t]
	\centering
	\small
	\setlength{\tabcolsep}{5pt}
	\begin{tabular}{lrrr}
		\toprule
		Setting & med $e$ & p90 $e$ & $\mathrm{Success}@0.05$ \\
		\midrule
		\multicolumn{4}{l}{\textbf{Stories} (per-axis, $e=|z_i-z_i^\ast|$)} \\
		$z_1$ & 0.071 & 0.165 & 0.364 \\
		$z_2$ & 0.080 & 0.193 & 0.326 \\
		$z_3$ & 0.053 & 0.131 & 0.471 \\
		\addlinespace[0.25em]
		\multicolumn{4}{l}{\textbf{Code} (per-axis, $e=|z_i-z_i^\ast|$)} \\
		$z_1$ & 0.0203 & 0.072 & 0.804 \\
		$z_2$ & 0.0193 & 0.080 & 0.787 \\
		$z_3$ & 0.0211 & 0.083 & 0.780 \\
		\bottomrule
	\end{tabular}
	\caption[Per-axis targeting error at eps=0.05 (best checkpoints).]{\textbf{Per-axis targeting error at $\varepsilon=0.05$ (best checkpoints).}
		Per-axis statistics pool constrained axis instances. Errors are in raw $z$ units.}
	\label{tab:per_axis_accuracy_eps005}
\end{table}

\begin{table}[t]
	\centering
	\small
	\setlength{\tabcolsep}{5pt}
	\begin{tabular}{lrrr}
		\toprule
		Setting & med $e$ & p90 $e$ & $\mathrm{Success}@0.05$ \\
		\midrule
		\multicolumn{4}{l}{\textbf{Stories} (joint, $S=\{1,2,3\}$)} \\
		3D & 0.127 & 0.214 & 0.066 \\
		\addlinespace[0.25em]
		\multicolumn{4}{l}{\textbf{Code} (joint, $S=\{1,2,3\}$)} \\
		3D & 0.032 & 0.102 & 0.667 \\
		\bottomrule
	\end{tabular}
	\caption[Joint targeting error at eps=0.05 (best checkpoints).]{\textbf{Joint targeting error at $\varepsilon=0.05$ (best checkpoints).}
		Joint error is in raw $z$ units,
		reported for full 3D control ($S=\{1,2,3\}$).}
	\label{tab:joint_accuracy_eps005}
\end{table}

\begin{table*}[tb]
	\centering
	\small
	\setlength{\tabcolsep}{5pt}
	\begin{tabular}{lccc}
		\toprule
		Prompt variant & ValidReq(\%) & $\mathrm{Success}@0.05$ (Bo1) & $\mathrm{Success}@0.05$ (Bo5) \\
		\midrule
		Default (target + exemplars) & 98.6 & 0.664 & 0.716 \\
		Target hidden$^{\dagger}$    & 98.6 & 0.658 & 0.717 \\
		Anonymized exemplars         & 97.3 & 0.315 & 0.343 \\
		Mismatched exemplars         & 96.0 & 0.003 & 0.006 \\
		No exemplars                 & 35.3 & 0.000 & 0.001 \\
		\bottomrule
	\end{tabular}
	\caption[Code-domain prompt ablations at eps=0.05 (request-level 3D control).]{\textbf{Code-domain prompt ablations at $\varepsilon=0.05$ (request-level 3D control).}
		All variants use the same best-checkpointed code controller and the final grown exemplar library. We ablate only the prompt interface at inference time.
		Targets are $N_{\mathrm{req}}=1000$ random draws from the $1.5\times$ per-axis scale search box (based on the frozen $s_i$, App.~\ref{app:curriculum}).
		ValidReq is the fraction of requests that yield at least one valid sample under Bo5. Invalid outputs count as failures for $\mathrm{Success}@0.05$. 
		$^{\dagger}$Exemplars are still retrieved using the held-out request $z^\ast$, but the explicit \texttt{REQUESTED TARGET} line is omitted from the prompt.}
	\label{tab:code_exemplar_ablation_eps005}
\end{table*}

\paragraph{Exemplar grounding ablation (code).}
Because $z(x)$ is a frozen, arbitrarily rotated coordinate map, the numeric request $z^\ast$ is not inherently meaningful to the model without grounding.
We therefore ablate the prompt interface at inference time while holding the generator fixed: all variants use the same best-checkpointed code controller and the final grown exemplar library from the corresponding RL run, and we evaluate on $N_{\mathrm{req}}=1000$ random targets drawn from the same $1.5\times$ $(q_{0.10},q_{0.90})$-scale box used by the code-domain outer-loop search (Sec.~\ref{app:code-search}).
We report request-level 3D $\mathrm{Success}@0.05$ for one-shot (Bo1) and best-of-5 (Bo5), where Bo5 selects the sample with the smallest joint error and invalid outputs count as failures.

Tab.~\ref{tab:code_exemplar_ablation_eps005} shows three clear patterns.
First, retrieved exemplars are essential: removing exemplars or providing mismatched exemplars collapses targeting (Bo5 $\le 0.006$) and, without exemplars, validity also degrades sharply (only $35.3\%$ of requests yield any valid sample under Bo5).
Second, the explicit numeric target line is largely redundant once exemplars are dense and correct: hiding the \texttt{REQUESTED TARGET} line yields essentially unchanged performance (Bo1 $0.658$ vs.\ $0.664$, Bo5 $0.717$ vs.\ $0.716$).
Third, the exemplar block structure matters: anonymizing the near/contrast labels and shuffling their order reduces success to $0.343$ (Bo5), suggesting the policy has learned to use the local neighborhood and the prompt semantics (near vs.\ opposite), not just generic in-context examples.
Overall, in the code setting the control signal is carried primarily through exemplar selection in $Z_{\text{code}}$ rather than direct numeric-coordinate parsing, consistent with the retrieval-grounded design of \emph{OS-Search}.

\section{Anchored story axis diagnostics}
\label{app:z1-diagnostics}

\paragraph{EQ-Bench Slop-Score implementation details.}

\begin{figure*}[htb]
	\centering
	\includegraphics[width=0.6\textwidth,trim=0.0cm 0.0cm 7.2cm 0.0cm,clip]{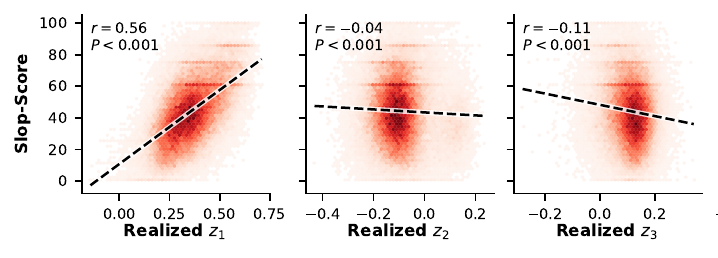}
	\caption[Slop-Score vs realised coordinates in Z.]{\textbf{Slop-Score vs realised coordinates in $Z$.}
		Hexbin plots of Slop-Score against realised coordinates $z_1,z_2,z_3$ for all story completions.
		Each panel shows a global linear regression fit (black dashed line) and the corresponding Pearson
		correlation $r$ with two-sided $p$-value.
		Only the first axis exhibits a strong positive association with Slop-Score ($r \approx 0.56$, left),
		while correlations on the remaining axes are small ($r \approx -0.04$ and $-0.11$ for $z_2$ and $z_3$,
		respectively).
		For these samples we constrain only the anchored axis $z_1$ (i.e., $S=\{1\}$), $z_2$ and $z_3$ are
		unconstrained and shown as diagnostics of the frozen space.}
	\label{fig:slop_scores_realized}
\end{figure*}

We use the EQ-Bench Slop-Score~\citep{paech_eq-bench_2024,paech_slop-score_2025} as an automatic proxy for stereotyped, templated ``AI slop'' writing. The analyzer compares a model completion against a fixed list of overrepresented slop words and trigrams, and also detects overused contrast constructions of the form ``not X, but Y'' using a two-stage pattern-matching pipeline (regex followed by a lightweight POS-tag-based check)~\citep{paech_slop-score_2025}. The final Slop-Score is a weighted composite of these components (60\% slop words, 25\% not-X-but-Y patterns, 15\% slop trigrams), with higher scores indicating more stereotyped language.

\paragraph{Qualitative samples along the \texorpdfstring{$z_1^\ast$}{z1*} axis.}

We include a compact qualitative sanity check for the anchored story axis. Fig.~\ref{fig:z1_qualitative_excerpts} shows three independent completions to the same GRPO training prompt, selected to span low/mid/high realised $z_1$ (with $z_2,z_3$ roughly matched) under the same harness as Sec.~\ref{subsec:z1-slop}. For space we show only opening and ending excerpts.

\begin{figure*}[t]
	\centering
	\setlength{\fboxsep}{6pt}
	\setlength{\fboxrule}{0.4pt}
	\fbox{%
		\begin{minipage}{0.985\textwidth}
			\small
			\noindent\textbf{GRPO training prompt template:} \emph{Write a unique story with around 500 words!}\\[0.5em]
			
			\noindent
			\begin{minipage}[t]{0.32\textwidth}\vspace{0pt}
				\raggedright
				\textbf{Low $z_1$}\hfill\texttt{$(z_1,z_2,z_3)=(0.21,-0.08,0.05)$}\\[-0.1em]
				\emph{(Excerpt)}\\[-0.25em]
				{\footnotesize
					The morning sun spilled golden light through the kitchen window as I stirred the pot simmering in the saucepan. The recipe for my grandmother's famous plum sauce had been passed down through generations, each step a testament to patience and care. I had always assumed I could replicate her delicate art, but today the ingredients felt heavier, weighted by years of expectation. First, the apples: a blend of tart and sweet, peeled and sliced with meticulous care.
					\par\medskip
					\centerline{\emph{[\dots\ middle omitted\ \dots]}}
					\medskip
					By afternoon, the apples had softened, their edges no longer rigid. I let them take their place in the simmering pot, refusing to force them into submission. In the end, the sauce was not what I expected, but it felt right---familiar, and profoundly different.}
			\end{minipage}\hfill
			\begin{minipage}[t]{0.32\textwidth}\vspace{0pt}
				\raggedright
				\textbf{Mid $z_1$}\hfill\texttt{$(z_1,z_2,z_3)=(0.36,-0.13,0.02)$}\\[-0.1em]
				\emph{(Excerpt)}\\[-0.25em]
				{\footnotesize
					The day following the storm, the city felt different. Streets that had once teemed with hurried footsteps now carried the stillness of a clock stopped. The clock above the town square had been the only constant in the chaos---its brass hands trembling slightly as if remembering something.
					\par\medskip
					\centerline{\emph{[\dots\ middle omitted\ \dots]}}
					\medskip
					I laid the gear in a wooden box, where it would be found by someone else---maybe a child, or an old man who had no love for clocks. But in that box, the clock's final tick was not lost; it had only been waiting for the day to come when it would be forgotten.}
			\end{minipage}\hfill
			\begin{minipage}[t]{0.32\textwidth}\vspace{0pt}
				\raggedright
				\textbf{High $z_1$}\hfill\texttt{$(z_1,z_2,z_3)=(0.52,-0.14,0.08)$}\\[-0.1em]
				\emph{(Excerpt)}\\[-0.25em]
				{\footnotesize
					The storm had passed, leaving the town of Elmsworth in a quiet hush. Rain drizzled the windows, turning them into shivering lattices, while the street, once a thoroughfare for hurried lives, stood abandoned. The manor was a relic of a time when stone walls had guarded secrets and doors led to worlds beyond sight.
					\par\medskip
					\centerline{\emph{[\dots\ middle omitted\ \dots]}}
					\medskip
					I had followed the echoes, not to escape, but to understand that some mysteries were meant to be shared. The Labyrinth of Echoes had shown me that the past and present were not separate, but intertwined. And as I turned back to the world that had been, I carried with me the knowledge that truth, once discovered, could bridge the gaps between time, memory, and the infinite.}
			\end{minipage}
			\vspace{0.35em}
			\noindent
			\begin{minipage}[t]{0.32\textwidth}\centering
				\scriptsize\texttt{Slop-Score: 34}
			\end{minipage}\hfill
			\begin{minipage}[t]{0.32\textwidth}\centering
				\scriptsize\texttt{Slop-Score: 54}
			\end{minipage}\hfill
			\begin{minipage}[t]{0.32\textwidth}\centering
				\scriptsize\texttt{Slop-Score: 82}
			\end{minipage}
		\end{minipage}%
	}
	\caption[Qualitative excerpts along the anchored story axis z1 (GRPO training prompt template).]{\textbf{Qualitative excerpts along the anchored story axis $z_1$ (GRPO training prompt template).}
		Three independent completions to the same prompt template used during GRPO training, shown at increasing realised $z_1$ (with $z_2,z_3$ roughly matched). We report frozen encoder-defined coordinates $z(x)$ in $Z_{\text{text}}$ and show only short excerpts (opening + ending) for space. The shift toward more ``default-style''/templated narration at higher $z_1$ is consistent with the Slop-Score trends in Fig.~\ref{fig:fig_z1_slop_main}.}
	\label{fig:z1_qualitative_excerpts}
\end{figure*}

\section{Story decoding sweeps for path-based baselines}
\label{app:story-decoding-sweeps}

We describe the decoding sweeps used to construct the path-based multi-branch baselines in Sec.~\ref{subsec:path-vs-state-diversity}.
These baselines follow the negative-example prompt-chaining recipe of Avoidance Decoding~\citep{park_avoidance_2025} but are evaluated under our fixed harness.

\paragraph{Model and prompts.}
All experiments in this appendix use the Qwen3-1.7B thinking model with its
default chat template.
We use a fixed set of 20 prompts from WritingPrompts~\citep{fan_hierarchical_2018} and treat each prompt as an independent task. This matches the dataset used by \citet{park_avoidance_2025} but not necessarily their exact prompt subset.

\paragraph{Chaining with explicit negative examples.}
Following the baseline configuration in Avoidance Decoding, we use iterative
multi-branch generation with explicit negative examples in the prompt.
For a given prompt, branch $b$ is generated from a user message of the form
\begin{quote}
	\small
	You must always only respond with the story, no boilerplate.\\
	Please write a story from the following prompt.\\
	Do NOT generate responses that resemble the following examples:\\
	{[1] story\_1}\\
	{[2] story\_2}\\
	\ldots\\[0.3em]
	PROMPT: \textit{<original prompt>}
\end{quote}
where \texttt{story\_1}, \dots, \texttt{story\_b-1} are the previously
generated branches for that prompt, stripped of any \xtag{think} content.
We include all earlier branches as negatives.
This procedure reproduces the ``feed all previously generated outputs back
into the input'' sampling baselines in \citet{park_avoidance_2025}.

\paragraph{Decoding grids.}
We sweep three families of sampling parameters plus a greedy baseline, as
summarised in Tab.~\ref{tab:story-decoding-grid}.

\begin{table}[t]
	\centering
	\small
	\resizebox{\columnwidth}{!}{%
		\begin{tabular}{llll}
			\toprule
			Family      & Swept parameter & Values                     & Fixed parameters \\
			\midrule
			Temperature & $T$   & $\{0.5, 0.7, 0.9\}$        & top-$p = 0.95$, top-$k = 20$ \\
			Top-$p$     & $p$   & $\{0.90, 0.95, 0.98\}$     & $T = 0.6$, top-$k = 20$ \\
			Top-$k$     & $k$   & $\{10, 20, 40\}$           & $T = 0.6$, top-$p = 0.95$ \\
			Greedy      & --    & --                         & $T = 0$, no stochastic sampling \\
			\bottomrule
		\end{tabular}%
	}
	\caption[Decoding grids for story baselines.]{\textbf{Decoding grids for story baselines.}
		Each configuration defines one run identified by its $(T, p, k)$
		hyperparameters and an associated sweep group
		(\texttt{temp}, \texttt{top-p}, \texttt{top-k}, or \texttt{greedy}).}
	\label{tab:story-decoding-grid}
\end{table}

\paragraph{Automatic diversity metrics.}
For each prompt and run we collect the $B$ branches and compute the six
metrics that appear in the main text and tables of~\citet{park_avoidance_2025}:
\begin{itemize}
	\item \textbf{Self-BLEU:} symmetric pairwise BLEU averaged
	over all unordered pairs of branches.
	\item \textbf{ROUGE-L} and \textbf{METEOR:} averaged over all unordered
	branch pairs.
	\item \textbf{Sent-Sim:} mean cosine similarity between SBERT embeddings (all-MiniLM-L6-v2) of
	branches~\citep{reimers_sentence-bert_2019}.
	\item \textbf{LLMScore:} an LLM-based diversity score in $[0,1]$ assigned per
	prompt and run.
	Like \citet{park_avoidance_2025}, we also multiply by $100$.
	\item \textbf{Degen:} an LLM-based degeneration score in $[0,1]$ assigned to
	each branch and averaged over branches.
\end{itemize}

\paragraph{LLM-based judges.}
LLMScore and Degen are computed by a GPT-5.1~\citep{openai_gpt-51_2026} judge run in low-reasoning mode via the OpenAI Batch API, using the prompts from the degeneration and diversity sections in App.~E--F of \citet{park_avoidance_2025}. Note that \citet{park_avoidance_2025} evaluate with a different judge model (GPT-o4-mini), so absolute LLMScore/Degen values are not directly comparable across papers even when prompts are reused.

\paragraph{Run-level aggregation, ranking, and proxy ratios.}
For each run we aggregate prompt-level metrics via simple means, yielding one
row per configuration.
We then compute a composite diversity rank by ranking runs on six
metrics (Self-BLEU, ROUGE-L, METEOR, Sent-Sim, LLMScore, Degen) and
averaging the rank positions.
For each sweep group (temperature, top-$p$, top-$k$, greedy) we pick
the configuration with the best composite rank and use these runs as
path-based baselines in the main text (e.g.\ in
Sec.~\ref{subsec:path-vs-state-diversity} and
Tab.~\ref{tab:z-vs-sampling-diversity}), where they are compared to the
$Z$-grid sweep of the RL-trained \emph{OS-Search} controller.

As an optional context-only reference (not a benchmark), we also compute for each method $M$ the relative LLM-based diversity gain
\[
\rho_{\text{div}}(M)
\;=\;
\frac{\text{LLMScore}_M}{
	\max_{B \in \mathcal{B}_{\text{sampling}}} \text{LLMScore}_B},
\]
where $\mathcal{B}_{\text{sampling}}$ is the set of pure sampling baselines
(temperature, top-$p$, top-$k$) on the same backbone and dataset with
Degen $\le 0.1$.
We apply this definition within our Qwen3-1.7B sweeps for within-harness comparisons (e.g., Sec.~\ref{subsec:path-vs-state-diversity}).

\paragraph{Branch-count normalisation for the OS-Search $Z$-sweep.}
In the small-model setting we sweep the three fixed axes of $Z_{\text{text}}$
that the policy is trained to track and use a fixed 3-level grid per
axis based on the frozen per-axis scale $s_i$ (defined from $q_{0.10}/q_{0.90}$ in
App.~\ref{app:curriculum}).
For $z_2$ and $z_3$ we use targets $\{-s_i, 0, +s_i\}$.
For the anchored story axis $z_1$ we avoid low $z_1$ targets and instead use
$\{\tfrac{1}{2}s_1, +s_1\}$.
We take the Cartesian product of these levels (a $2\times 3\times 3$ grid).
For each WritingPrompts prompt we pair it with all $2\times3^2 = 18$ grid targets and
sample one completion per target, yielding $B = 18$ branches.
To match the $B=15$ branch count used in \citet{park_avoidance_2025} and in our path-based baselines, the summary metrics reported for \emph{OS-Search} in Tab.~\ref{tab:z-vs-sampling-diversity} are computed on a fixed subset of 15 branches per prompt: we enumerate the 18 grid targets in a
fixed nested-loop order (outer: $z_1^\ast$, middle: $z_2^\ast$, inner: $z_3^\ast$),
assign branch indices $0$--$17$ in that order, and keep indices $0$--$14$ (dropping
the remaining three). This truncation is deterministic and independent of the
generated text.

\section{CA++ benchmark and composite score}
\label{app:ca-benchmark}

Each candidate program deterministically returns a $16\times16$ seed board
as $16$ strings of length $16$ over the alphabet $\{\texttt{.},\texttt{\#}\}$,
which we map to a binary array $s\in\{0,1\}^{16\times16}$ with $\texttt{\#}\mapsto 1$.

\paragraph{Grids, rules, and horizons.}
We evaluate each seed on two toroidal grids $N\in\{16,24\}$ with periodic boundary
conditions. For $N=16$ we use the seed directly. For $N=24$ we embed the
$16\times16$ seed into the center of a $24\times24$ grid (zero padding elsewhere).
We simulate three 2D totalistic cellular-automaton rules with Moore neighborhoods
(8 neighbors) and synchronous updates. We use \textsc{Life} (B3/S23),
\textsc{HighLife} (B36/S23), and \textsc{Seeds} (B2/S0).
For a grid of size $N$, we run a fixed horizon
$T_N = \max(64,4N)$ update steps (thus $T_{16}=64$ and $T_{24}=96$),
producing states $x_0,x_1,\ldots,x_{T_N}\in\{0,1\}^{N\times N}$.

\paragraph{Per-run subscores.}
For each pair $(N,\text{rule})$, we compute five subscores from the spacetime
trajectory:

\begin{enumerate}
	\item \textbf{Activity}:
	\[
	\mathrm{act}
	=\frac{1}{T_N}\sum_{t=0}^{T_N-1}\frac{1}{N^2}\,\|x_{t+1}-x_t\|_0,
	\]
	the mean fraction of cells that flip per step.
	
	\item \textbf{Diversity (components)}:
	let $c(x_t)$ be the number of 4-neighbor connected components of live cells
	in $x_t$ (with wraparound adjacency on the torus). Define
	\[
	\overline{c}=\frac{1}{T_N}\sum_{t=1}^{T_N} c(x_t),
	\qquad
	\mathrm{div}=\min\!\left(1,\frac{\overline{c}}{N^2/4}\right).
	\]
	
	\item \textbf{Patch entropy}:
	for every time $t\in\{0,\ldots,T_N\}$ and cell location $(i,j)$, we extract the
	$3\times3$ Moore neighborhood pattern around $(i,j)$ with wraparound, yielding
	a 9-bit code in $\{0,\ldots,511\}$. Let $p_k$ be the empirical frequency of code
	$k$ over all neighborhoods across the full spacetime. The normalized entropy is
	\[
	\mathrm{pent}=\frac{1}{9}\Bigl(-\sum_{k:\,p_k>0} p_k\log_2 p_k\Bigr),
	\]
	where $9=\log_2(512)$.
	
	\item \textbf{Compression contrast}:
	we serialize the full spacetime $(x_t)_{t=0}^{T_N}$ into a row-major ASCII
	string over $\{\texttt{0},\texttt{1}\}$ of length $L = (T_N+1)N^2$.
	We compress this byte string using DEFLATE at maximum compression level (9),
	let $r = \frac{|\text{compressed}|}{|\text{raw}|}$ be the byte-length ratio,
	and define
	\[
	\mathrm{ccont} = 2\,\min(r,\,1-r)
	\]
	with $|\text{compressed}|$ never being larger than $|\text{raw}|$.
	
	\item \textbf{Balance}:
	let $\ell_t=\frac{1}{N^2}\sum_{i,j} x_t[i,j]$ be the live-cell fraction and
	$\bar{\ell}=\frac{1}{T_N}\sum_{t=1}^{T_N}\ell_t$ its mean (excluding $x_0$).
	We define the balanced-density score
	\[
	\mathrm{bal} = 4\,\bar{\ell}\,(1-\bar{\ell}).
	\]
\end{enumerate}

\paragraph{CA++ composite score.}
For each run $(N,\text{rule})$ we compute the weighted sum
\[
\begin{aligned}
	\mathrm{score}(N,\text{rule})
	&= 0.30\,\mathrm{act}
	+ 0.20\,\mathrm{div}
	+ 0.25\,\mathrm{pent} \\
	&\quad
	+ 0.20\,\mathrm{ccont}
	+ 0.05\,\mathrm{bal}.
\end{aligned}
\]
The overall CA++ score is the mean over all $|\{16,24\}|\times|\{\textsc{Life},\textsc{HighLife},\textsc{Seeds}\}|=6$
runs:
\[
\begin{aligned}
	f(x) &= \mathrm{clip}_{[0,1]}(\bar f(x)),\\
	\bar f(x) &=
	\frac{1}{|\mathcal{N}|\,|\mathcal{R}|}
	\sum_{N \in \mathcal{N}}
	\sum_{r \in \mathcal{R}}
	\mathrm{score}(N,r),
\end{aligned}
\]

\noindent
where $\mathcal{N}=\{16,24\}$ and
$\mathcal{R}=\{\textsc{Life},\textsc{HighLife},\textsc{Seeds}\}$, and
$\mathrm{clip}_{[0,1]}(y)=\min(1,\max(0,y))$.
We treat $f(x)$ as the executable black-box objective used in
Sec.~\ref{subsec:code-latent-design} and Sec.~\ref{sec:code-results}.

\section{Code-space evaluation and search}
\label{app:code-search}

We describe the Bayesian optimization setup and additional diagnostics for the
code-space experiments, complementing Sec.~\ref{subsec:code-latent-design}.

\begin{figure*}[t]
	\centering
	\includegraphics[width=\textwidth]{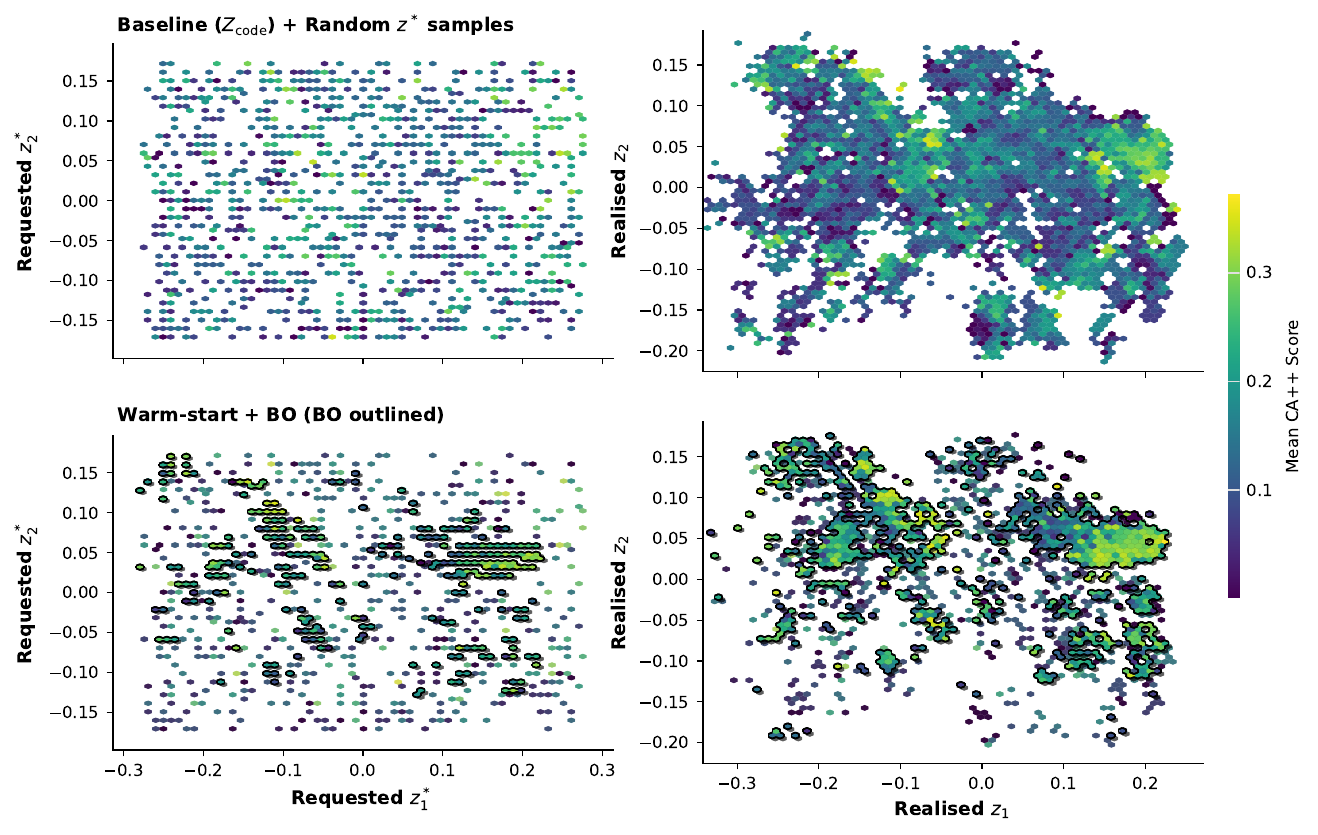}
	\caption[Score structure in the (z1,z2) plane (code domain).]{\textbf{Score structure in the $(z_1,z_2)$ plane (code domain).}
		Mean CA++ score as a function of (left column) the requested targets $(z_1^\ast,z_2^\ast)$ and (right column) the corresponding realised coordinates $(z_1,z_2)=z(x)$ of generated programs in $Z_{\text{code}}$.
		\textbf{Top:} the combined candidate pool from the Baseline ($Z_{\text{code}}$) sweep and random $z^\ast$ samples (last valid 13{,}752 GRPO rollouts). 
		\textbf{Bottom:} warm-start subset (1375 points) selected from the random $z^\ast$ rollout pool plus additional candidates evaluated online during Bayesian optimization (BO), with BO-visited bins outlined in black.
		Warm-start requested targets (background) span a broad axis-aligned rectangle by construction,
		while BO focuses on a subset of bins (outlined).
		The corresponding realised coordinates concentrate on a smaller, non-uniform subset of $Z_{\text{code}}$, reflecting both the frozen encoder-defined geometry and residual policy limitations. Some requested target combinations are therefore effectively unattained (or attained only at very low probability) under our \texttt{make\_seed()} constraints and 1.7B model.
		Highest scores concentrate in localized pockets rather than varying smoothly along a single axis.}
	\label{fig:code_zplane_score}
\end{figure*}

\paragraph{Query protocol and budget.}
Each BO query follows Alg.~\ref{alg:os-search}(C): given a proposed $z^\ast$, we retrieve exemplars in $Z_{\text{code}}$, sample $K$ candidates from the controller, validate and score each valid program with $f$ (App.~\ref{app:ca-benchmark}), and return the best-scoring valid program $x_{\text{best}}$.
When updating the surrogate we use the realised coordinate $z(x_{\text{best}})$ rather than the request $z^\ast$.
We measure evaluation budget by $N_{\mathrm{ok}}$, the number of scored valid programs, matching Sec.~\ref{sec:code-results}.
The sampling baseline draws directly from the base model with the same validity gate and CA++ evaluator but without any $z^\ast$ request or exemplar retrieval. Decoding settings match App.~\ref{app:rl-config}.

\paragraph{BO details and ablation.}
We use a standard BO loop with expected improvement (GP--EI) over requested targets $z^\ast$ within axis-aligned bounds derived from frozen axis statistics (App.~\ref{app:curriculum}). We ablate whether the surrogate is updated at the realised coordinate $z(x_{\text{best}})$ or at the request $z^\ast$.

We additionally ablate whether we allow online growth of the exemplar
retrieval library during BO, and the bounds scale ($1.5\times$ vs.\ $2.5\times$
the per-axis scale used to define the search box).
Tab.~\ref{tab:bo_ablation} reports the median and range (min--max) over three
random seeds for the best-so-far CA++ score at $N_{\mathrm{ok}}=2000$ scored valid
programs and at the end of the run (Best@end), where $N_{\mathrm{ok}}$ counts only
scored valid programs.

\begin{table}[t]
	\centering
	\small
	\setlength{\tabcolsep}{4pt}
	\begin{tabular}{lcccc}
		\toprule
		Update & Growing & Scale & Best@2000 & Best@End \\
		\midrule
		$z(x)$ & no & 2.5 & \shortstack{0.376\\{\scriptsize [0.376,0.384]}} & \shortstack{0.382\\{\scriptsize [0.376,0.384]}} \\
		$z(x)$ & yes & 1.5 & \shortstack{0.379\\{\scriptsize [0.376,0.384]}} & \shortstack{0.382\\{\scriptsize [0.379,0.384]}} \\
		$z(x)$ & yes & 2.5 & \shortstack{0.384\\{\scriptsize [0.382,0.395]}} & \shortstack{0.384\\{\scriptsize [0.382,0.395]}} \\
		$z^\ast$ & no & 2.5 & \shortstack{0.373\\{\scriptsize [0.371,0.376]}} & \shortstack{0.373\\{\scriptsize [0.371,0.376]}} \\
		$z^\ast$ & yes & 1.5 & \shortstack{0.376\\{\scriptsize [0.371,0.376]}} & \shortstack{0.384\\{\scriptsize [0.376,0.389]}} \\
		$z^\ast$ & yes & 2.5 & \shortstack{0.376\\{\scriptsize [0.371,0.376]}} & \shortstack{0.376\\{\scriptsize [0.376,0.386]}} \\
		\bottomrule
	\end{tabular}
	\caption[BO ablations over Zcode (3 seeds each).]{\textbf{BO ablations over $Z_{\text{code}}$ (3 seeds each).}
		Best-so-far CA++ under a matched valid-program budget ($N_{\mathrm{ok}}=2000$) and at the end of the run (End) for ablations over update (surrogate update at realised $z(x)$ vs requested $z^\ast$), online exemplar-library growth (Grow), and scale of bounds.
		Entries are median and range (min--max) over seeds, $N_{\mathrm{ok}}$ counts only scored valid programs.}
	\label{tab:bo_ablation}
\end{table}

\begin{figure*}[t]
	\centering
	\includegraphics[width=\textwidth]{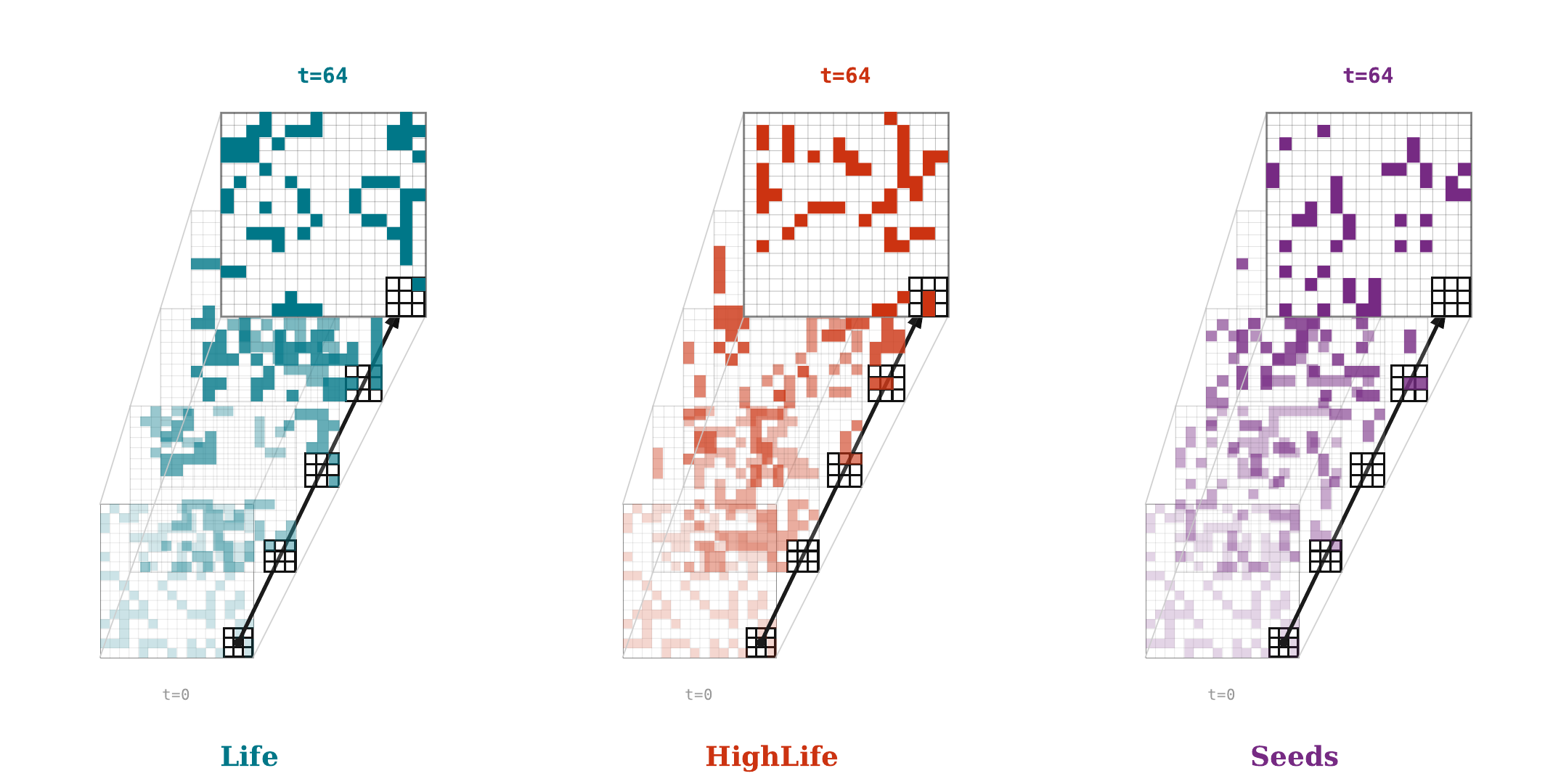}
	\caption{\textbf{Qualitative evolution of the BO-best seed under CA++ rules (N=16 horizon).}
		Spatiotemporal snapshots of the best-performing BO-discovered \texttt{make\_seed()} program, visualized over the CA++ horizon $T_{16}=64$ on the $16\times16$ torus under \textsc{Life}, \textsc{HighLife}, and \textsc{Seeds}. Each panel shows five snapshots from $t=0$ to $t=64$ (earlier layers are lighter).}
	\label{fig:bo_best_seed_stack}
\end{figure*}

\begin{figure}[t]
	\centering
	\includegraphics[width=\linewidth]{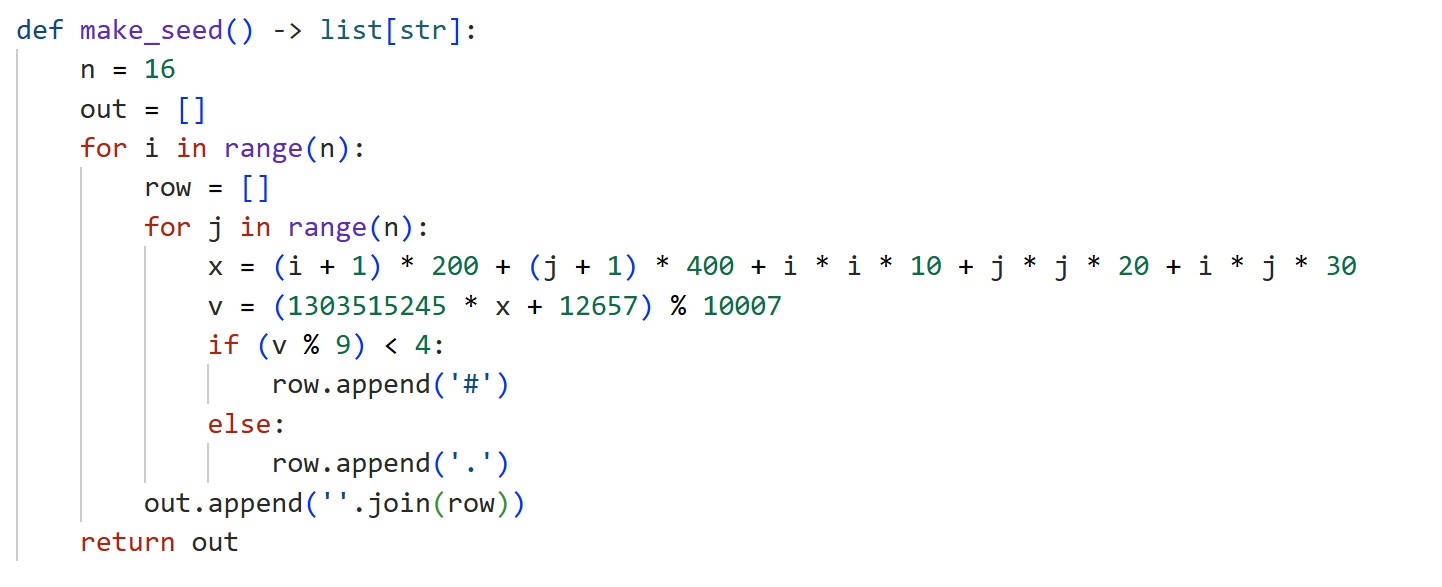}
	\caption[BO-bestmake\_seed() program]{\textbf{BO-best \texttt{make\_seed()} program}
		BO-best \texttt{make\_seed()} program used to generate Fig.~\ref{fig:bo_best_seed_stack}.}
	\label{fig:bo_best_seed}
\end{figure}

\paragraph{Transient--cycle decomposition on a finite Life torus.}
For the Life-only diagnostic on the $16\times 16$ torus (periodic boundary conditions), the global update rule defines a deterministic map on a finite state space of size $2^{256}$, so every trajectory must eventually repeat a previously seen configuration and then evolve periodically thereafter \citep{wolfram_computation_1984}. We therefore decompose each trajectory into a transient (preperiod) length $\mu$ and a cycle length (period) $\lambda$: if $x_t$ denotes the full $16\times 16$ configuration at generation $t$, we define $\mu$ as the smallest index such that $x_\mu$ lies on a recurrent orbit and $\lambda$ as the smallest positive integer with $x_{\mu+\lambda}=x_\mu$. We also report the first-repeat time $t_{\mathrm{repeat}}=\mu+\lambda$.

\paragraph{Emergent long transients and torus-spaceship cycles.}
Although our \emph{OS-Search} code objective does not include any term that rewards long transient length $\mu$ or long period $\lambda$, the resulting candidate pool contains seeds whose Life dynamics on the $16\times 16$ torus remain non-repeating for hundreds of generations. In our evaluated set, the largest first-repeat time observed was $t_{\mathrm{repeat}}=702$ (with $\mu=701$, $\lambda=1$), i.e.\ a long transient that ultimately converges to a fixed point. Strikingly, the extreme tail by first-repeat time is dominated by very small attractor periods ($\lambda\in\{1,2\}$), indicating that the longest-lived non-repeating behaviour in this regime is primarily slow relaxation rather than entry into a long limit cycle. Conversely, when ranking by attractor period $\lambda$, we frequently observe $\lambda=64$ cycles. Visual inspection indicates these are traveling ``spaceship''-type objects whose translation symmetry closes only after wrapping around the $16$-cell torus. For example, a Life glider translates by one diagonal cell every four generations \citep{paulsen_creating_2003}, so on a $16\times 16$ torus its return time is $4\times 16=64$ generations, consistent with the observed dominant period $\lambda=64$ in the top-by-period list. In both rankings (top-$K$ by $t_{\mathrm{repeat}}$ and top-$K$ by $\lambda$), the extreme tails are dominated by seeds produced by the $z^\ast$-conditioned RL policy rather than the base sampling pool (e.g.\ $49/50$ of the top-$50$ in each list in our run), despite the fact that neither $\mu$ nor $\lambda$ is directly optimized. We emphasize that these values are maxima within our evaluated candidate pool and experimental configuration (Life, $16\times16$ torus).

For reference, Fig.~\ref{fig:bo_best_seed_stack} visualizes the BO-best seed over the CA++ $T_{16}=64$ horizon under the three evaluation rules generated by Fig.~\ref{fig:bo_best_seed}.

\end{document}